\pdfoutput=1

\documentclass[11pt]{article}
\PassOptionsToPackage{table}{xcolor}

\usepackage[preprint]{acl}

\usepackage{times}
\usepackage{latexsym}

\usepackage[T1]{fontenc}

\usepackage[utf8]{inputenc}

\usepackage{microtype}

\IfFileExists{inconsolata.sty}{\usepackage{inconsolata}}{}

\usepackage{colortbl}
\usepackage{multirow}
\usepackage{booktabs}
\usepackage{subcaption}
\usepackage{adjustbox}

\usepackage{graphicx}
\usepackage{amsmath,amssymb,amsfonts}
\usepackage[utf8]{inputenc}
\usepackage[T1]{fontenc}
\usepackage{listings}
\usepackage{mdframed}
\usepackage{geometry}

\usepackage{algorithm}
\usepackage{algorithmicx} 
\usepackage{algpseudocode} 
\usepackage{makecell}
\usepackage{multirow}

\usepackage{times}
\usepackage{latexsym}
\usepackage{microtype}
\IfFileExists{inconsolata.sty}{\usepackage{inconsolata}}{}
\usepackage{hyperref}
\usepackage{tcolorbox}
\usepackage{enumitem}
\usepackage{dblfloatfix} 

\definecolor{systemColor}{RGB}{70,130,180}
\definecolor{driverColor}{RGB}{46,139,87}
\definecolor{navigatorColor}{RGB}{205,92,92}
\definecolor{fixColor}{RGB}{128,64,0}  
\definecolor{reviewColor}{RGB}{128,0,0}  

\definecolor{lightgreen}{RGB}{144,238,144}
\definecolor{medgreen}{RGB}{100,200,100}
\definecolor{darkgreen}{RGB}{50,180,50}
\definecolor{lightyellow}{RGB}{255,255,200}
\definecolor{lightpeach}{RGB}{255,218,185}
\definecolor{lightpink}{RGB}{255,182,193}

\newenvironment{messagebox}[2]{
    \begin{mdframed}[
        linecolor=#1,
        linewidth=2pt,
        backgroundcolor=#1!10,
        roundcorner=10pt,
        skipabove=10pt,
        skipbelow=10pt,
        nobreak=true,  
        splitbottomskip=0pt,
        splittopskip=0pt,
        innertopmargin=0pt,
        innerbottommargin=0pt
    ]
    \textcolor{#1}{\textbf{#2}}
    \par\medskip
}{
    \end{mdframed}
}

\author{
\mdseries
Junhao Chen\textsuperscript{1} \quad
Xiang Li\textsuperscript{2} \quad
Mingjin Chen\textsuperscript{3} \quad
Boran Zhang\textsuperscript{4} \quad
Henghaofan Zhang\textsuperscript{5} \\
Yibin Xu\textsuperscript{6} \quad
Yuehan Cui\textsuperscript{7} \quad
Fangsheng Weng\textsuperscript{8} \quad
Fei Ma\textsuperscript{9} \\
Qi Tian\textsuperscript{9} \quad
Ruqi Huang\textsuperscript{1} \quad
Hao Zhao\textsuperscript{1,10} \\
\\
\textsuperscript{1}THU \quad
\textsuperscript{2}PKU \quad
\textsuperscript{3}PolyU, Hong Kong \quad
\textsuperscript{4}USTC \quad
\textsuperscript{5}UESTC \quad
\textsuperscript{6}Tongji University \\
\textsuperscript{7}Tianjin University \quad
\textsuperscript{8}Independent Researcher \quad
\textsuperscript{9}Guangming Lab \quad
\textsuperscript{10}BAAI \\
\rule{0pt}{0.3ex}\\
\normalsize Project page: \href{http://yisuanwang.github.io/PairCoder}{\texttt{http://yisuanwang.github.io/PairCoder}}\\
\rule{0pt}{0.1ex}
}

\title{PairCoder++: Pair Programming as a Universal Paradigm for Verified\\ Code-Driven Multimodal and Structured-Artifact Generation}

\makeatletter
\AtBeginDocument{%
\g@addto@macro\@maketitle{%
  \vspace{-1.0cm}%
  \centering
  \includegraphics[width=0.98\textwidth]{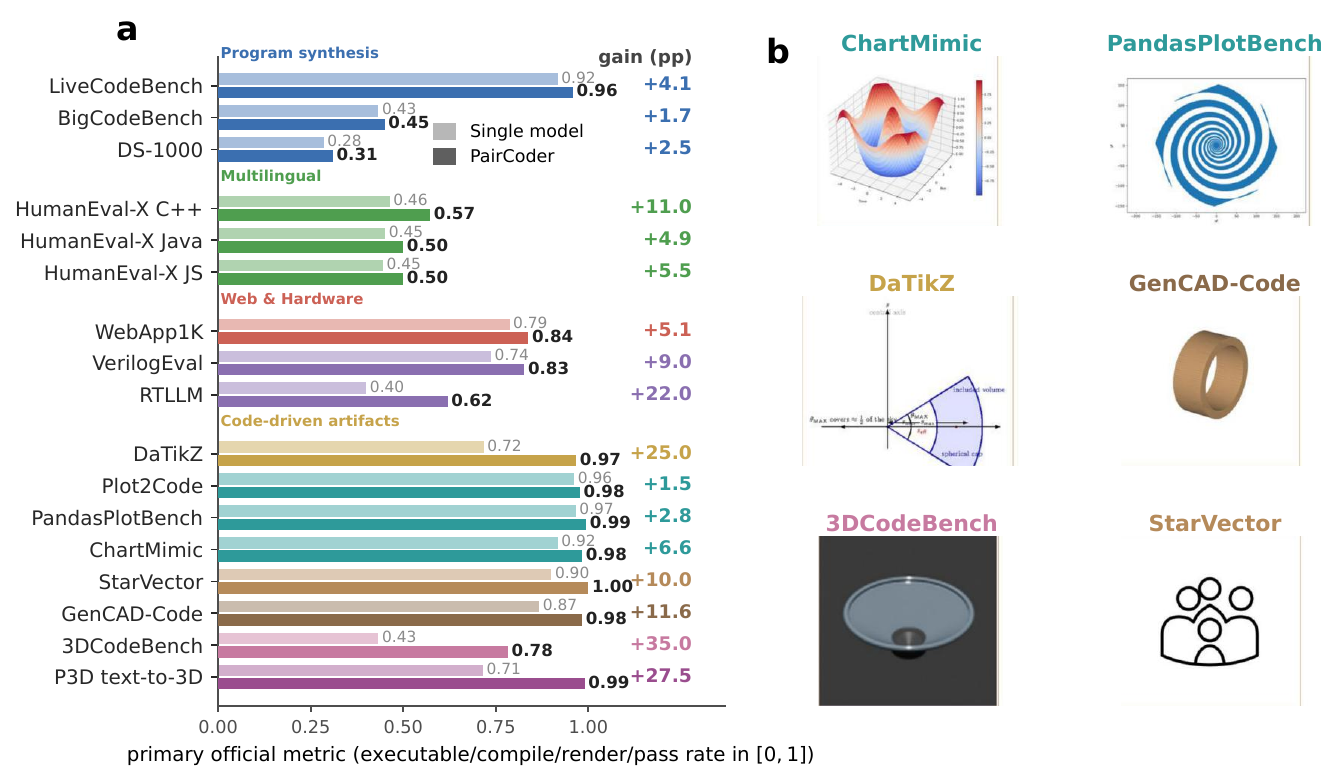}\par
  \vskip 0.4ex%
  \captionof{figure}{PairCoder improves code generation across every artifact domain, not only Python
    program synthesis. \textbf{(a)} All 17 public benchmarks at \texttt{gpt-5.4} (thinking off): the
    light bar is single-model, the dark bar (bold value) is PairCoder, coloured by benchmark family,
    with the absolute gain (pp) at right; each bar is the benchmark's primary official metric, an
    executable/compile/render/pass rate in $[0,1]$.
    P3D-Bench is shown on its text-to-3D track (400 cases); its image-to-3D and assembly-3D tracks
    are reported as a 3-case demonstration in the appendix (Table~\ref{tab:p3d_demo}).
    \textbf{(b)} Representative artifacts across six modalities: a chart (ChartMimic), a scientific
    figure (PandasPlotBench), a diagram (DaTikZ), a CAD model (GenCAD-Code), a 3D scene (3DCodeBench),
    and a vector graphic (StarVector).}%
  \label{fig:placeholder}%
  \vskip 0.4ex%
}%
}
\makeatother

\begin{document}
\maketitle

\begin{abstract}
Code is the medium through which large language models generate structured artifacts: charts,
scientific figures, vector graphics, CAD models, 3D scenes, and hardware designs are all
produced by writing programs. In this regime single pass inference is brittle, because the
compiler, renderer, or simulator that decides whether the artifact exists is invisible to the
model. We present \textbf{PairCoder}, which grounds review in the toolchain and realizes it as two agent pair programming: a Driver agent writes the program, a Navigator agent
reviews it against verification evidence (diagnostics, execution results, and renderings of
the current artifact beside the target), and the two switch roles when errors persist. Across
17 public benchmarks and seven models from three vendors, PairCoder improves essentially
every benchmark whose artifact is verifiable, on full official metric suites rather than
execution alone (for example, Blender scene executability 0.20 to 0.78; Ti\textit{k}Z compile
rate up 10 to 30 points on every model), at 2.9 to 9.2 times single model cost (about 7 times overall). The improvements concentrate
where the toolchain provides an informative oracle and the baseline leaves headroom, and the
method ties or mildly regresses where the oracle is weak; we frame pair programming as a
reliable recipe for verified code driven generation.

\end{abstract}

\section{Introduction}

Code generation for structured data and multimodal content asks a model to write a program
whose value lies in the artifact it renders: scientific figures as Ti\textit{k}Z, charts as
matplotlib, icons as SVG, mechanical parts as CAD programs, 3D scenes as Blender scripts, web
applications as React projects, and circuits as Verilog
RTL~\cite{belouadi2024automatikz,yang2024chartmimic,wu2024plot2code,galimzyanov2024drawing,rodriguez2024starvector,alam2024gencad,gao20263dcodebench,cui2024webapp1k,iwbench,liu2023verilogeval,lu2024rtllm}.
The task is broadly useful and hard in a specific way: the program must survive an external
toolchain (a compiler, renderer, or simulator) that the model cannot observe while writing,
and the artifact must then match the target in structure, semantics, and geometry. Even
strong LLMs~\cite{openai-gpt-turbo,qwen,deepseekai2024deepseekv3technicalreport,li2023starcoder}
routinely emit programs that do not compile, do not run, or render something far from the
request~\cite{li2023cctest,zhang2025llm,tian2025codehalu}.

Existing remedies fall into two camps. One scales up collaboration: team style multi agent
frameworks~\cite{hong2023metagpt,qian2023communicative,li2023camel,wu2023autogen} and role
specialized pipelines~\cite{huang2023agentcoder,dong2024self,islam2024mapcoder,chen2023agentverse}
improve correctness but pay up to an order of magnitude token overhead, and recent agentic
creation systems show the same closed loop instinct in single
domains~\cite{pfaff2026scenesmith,chen2025idea23d}. The other camp keeps one model and adds
self correction from runtime
feedback~\cite{shinn2024reflexion,chen2023selfdebug,madaan2023selfrefine,gou2024critic},
which is cheap but inherits the blind spots of a single perspective. Neither camp offers a
domain general mechanism that anchors an independent reviewer to the toolchain, which is
exactly where code driven creation fails.

We argue that this task needs the smallest collaborative structure in which an independent
reviewer is grounded in verification, and human software engineering already has it: pair
programming~\cite{umapathy2017meta,lui2006pair,hannay2009effectiveness}. We propose
\textbf{PairCoder}, a two agent framework whose core idea is to turn code driven generation
into a toolchain verified dialogue: one agent writes the program, the other reviews it
against concrete verification evidence, and control of the keyboard follows the diagnosis.

PairCoder instantiates this idea with three components. (1) Driver and Navigator roles with
Self-Mirror prompting: the Driver generates and revises while the Navigator reviews, each
under a role specific system prompt over the shared task and memory, preserving two genuinely
different perspectives on the same code. (2) Verification grounded review: each candidate is
compiled, executed, or rendered by the benchmark's own toolchain, and the evidence, including
diagnostics, test outcomes, and a rendering of the current artifact beside the target, is
attached to the review; the Navigator must cite a concrete error or accept with
\texttt{[NOERROR]}, which prevents churn on already correct programs. (3) Error triggered
role switching with a quality fallback: persistent errors swap the roles so the agent that
diagnosed the fault repairs it, and an exhausted budget returns the candidate with the best
verified quality rather than the last one.

Across 17 public benchmarks spanning program synthesis, multilingual code, data science, web,
hardware, and six artifact modalities, under seven models from three vendors, PairCoder
improves essentially every benchmark whose artifact is verifiable (Figure~\ref{fig:placeholder}),
on full official metric suites rather than execution alone: Blender scene executability rises from 0.20 to 0.78,
Ti\textit{k}Z compile rate gains 10 to 30 points on every model, and chart, SVG, and CAD
benchmarks improve in execution rate, SSIM, and CLIP, while on classical synthesis PairCoder
stays competitive and often improves further (e.g.\ LiveCodeBench pass@1 from 0.94 to 0.99 at
\texttt{gpt-5.4-mini}) at 2.9 to 9.2 times the single pass cost. A consistent pattern emerges:
gains concentrate where the toolchain gives an informative oracle and the baseline leaves
headroom, with ties at saturated oracles and mild regressions where the oracle is weak.

Our contributions are threefold: (1) PairCoder, a two agent framework that grounds pair
programming in toolchain verification, unifying verification grounded review and error
triggered role switching into one protocol across modalities; (2) the first systematic
evidence that pair programming transfers from Python synthesis to structured data and
multimodal artifact generation, improving executability, visual similarity (SSIM, CLIP, DINO,
SigLIP-2), and geometry (Chamfer) across 17 benchmarks, seven models, and three vendors; and
(3) a qualitative heuristic that gains scale with oracle informativeness and baseline headroom,
characterizing when the collaboration helps, ties, or is wasted.

\section{Related Work}

\subsection{Large Language Models}
LLMs~\cite{openai2023gpt4,touvron2023llama,qwen,hui2024qwen2,yang2025qwen3,deepseekai2024deepseekv3technicalreport,deepseekai2025deepseekr1incentivizingreasoningcapability}
have made code one of their most impactful application
domains~\cite{chen2021evaluating,liu2024your,jiang2024survey}, supported by code specialized
model families~\cite{nijkamp2023codegen2,zheng2023codegeex,li2023starcoder,lozhkov2024starcoder2,roziere2023code}
and increasingly demanding
benchmarks~\cite{jain2024livecodebench,zhuo2024bigcodebench,lai2023ds1000,jimenez2024swebench,zhujiu,llmspark,iwbench}.
Foundation models also reach into multimodal perception, multimodal generation, and
robustness~\cite{chen2023towards,miao2026framessequencestemporallyconsistent,ye2024mmad,chen2023soulstyler,chen2026ovow},
yet single pass inference remains prone to logical errors and
hallucinations~\cite{zhang2025llm,tian2025codehalu,ji2023survey}. PairCoder is model
agnostic: any of these models can play Driver or Navigator, and the collaboration adds
reliability on top of whatever base capability the chosen model provides.

\subsection{Structured Data Generation}
A fast growing line of work, recently surveyed under the banner of multimodal code
intelligence, treats programs as the medium for generating
structured and multimodal content, where the value of the code lies in the artifact it renders:
scientific
figures~\cite{belouadi2024automatikz,belouadi2024detikzify}, charts and
plots~\cite{yang2024chartmimic,wu2024plot2code,galimzyanov2024drawing,yang2024matplotagent,zhao2025chartcoder},
vector graphics and animation~\cite{rodriguez2024starvector,wu2024chat2svg,chen2026lottiegpttokenizingvectoranimation,qiu2024sgpbench},
garment patterns~\cite{weng2026garmentgpt}, parametric
CAD~\cite{alam2024gencad,alrashedy2024cadcode}, procedural 3D
scenes~\cite{hu2024scenecraft,sun20233dgpt,gao20263dcodebench}, front end
code~\cite{si2024design2code,cui2024webapp1k}, and hardware
RTL~\cite{liu2023verilogeval,lu2024rtllm}. These benchmarks score the rendered artifact with
perceptual and geometric metrics~\cite{wang2004image,radford2021learning,oquab2023dinov2,tschannen2025siglip},
yet the methods they evaluate are almost exclusively single pass generators. A parallel line
of work synthesizes multimodal content directly with learned models rather than programs,
spanning 3D editing, texture, rigging, skeleton, and human centric video
generation~\cite{weng2026feedforward3deditinglearns,chen2026ultraman,sun2025drive,Sun_2026_CVPR,chen2026hvg3dbridgingrealsimulation,chen2026dancetogether};
PairCoder instead keeps the program as the medium, which makes the toolchain itself available
as a verifier. PairCoder targets exactly this family of tasks and improves the full official
metric suites rather than execution success alone; it is a concrete instance of the
verification centered multimodal code intelligence that recent surveys call for.

\subsection{Multi-Agent Collaboration}
Two strands of work attack the brittleness of single pass generation. Multi agent systems
divide labor across roles, from simulated teams and conversation
scaffolds~\cite{hong2023metagpt,qian2023communicative,li2023camel,wu2023autogen} to role
specialized pipelines~\cite{huang2023agentcoder,dong2024self,islam2024mapcoder,chen2023agentverse}
and domain specific agentic
creation~\cite{pfaff2026scenesmith,chen2025idea23d}; the other closes the loop with external
signals, through self correction and tool grounded
critique~\cite{shinn2024reflexion,chen2023selfdebug,madaan2023selfrefine,gou2024critic},
learned verifiers and repair loops~\cite{ni2023lever,tsai2024rtlfixer}, and agentic software
systems~\cite{yang2024sweagent,wang2024openhands}. Team scale systems pay an order of
magnitude token overhead and the domain built ones are engineered per task, while single
model self correction inherits its own blind spots. PairCoder occupies the intersection they
leave open: the smallest collaborative unit, two agents in the Driver and Navigator pattern
of human pair
programming~\cite{umapathy2017meta,lui2006pair,hannay2009effectiveness,williams2000strengthening},
with toolchain grounded review and error triggered role switching, applied as one uniform
protocol from program synthesis to code driven multimodal generation.

\section{Methods}

\begin{figure*}[tp]
    \centering
    \includegraphics[width=\textwidth]{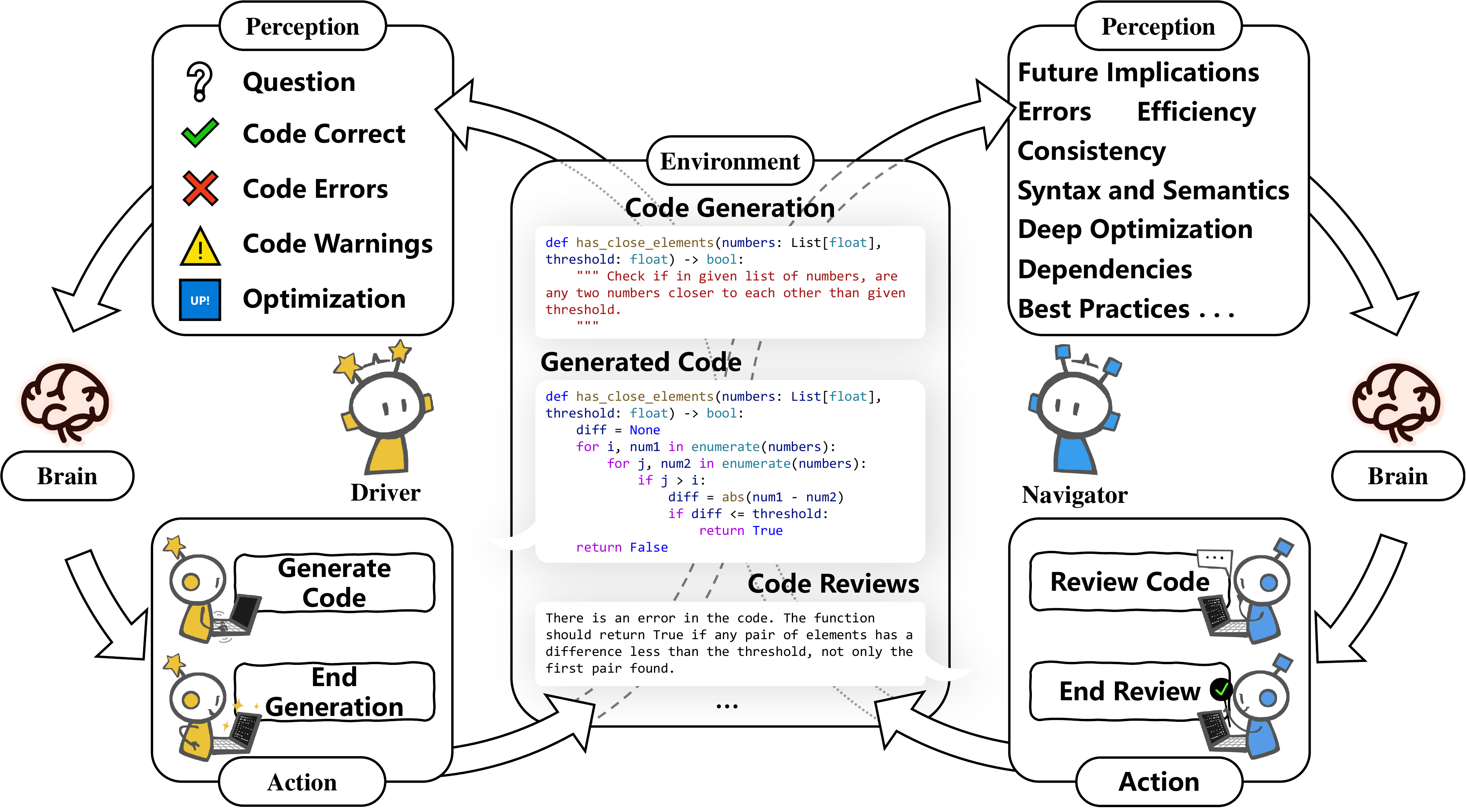}
    \vspace{-2mm}
    \caption{The PairCoder framework: the Driver generates and revises code, the Navigator reviews it against verification evidence, and persistent errors switch the roles.}
    \label{agent_framework}
\end{figure*}

\subsection{Overview}
The PairCoder framework, illustrated in Fig.~\ref{agent_framework}, adapts pair programming to automated code generation with Large Language Models (LLMs). Two agents alternate between proposal and review. At each iteration, the Driver produces a candidate solution and the Navigator critiques it, after which the Driver revises the code. This interaction continues until the solution is accepted or the iteration budget is exhausted.

Formally, given a programming task $q \in \mathcal{Q}$, PairCoder seeks to generate code $c^* \in \mathcal{C}$ that maximizes both correctness and quality:
\begin{equation}
c^* = \arg\max_{c \in \mathcal{C}} P(c \mid q) \cdot \text{Quality}(c, q)
\end{equation}

\begin{figure*}[tp]
    \centering
    \includegraphics[width=\textwidth]{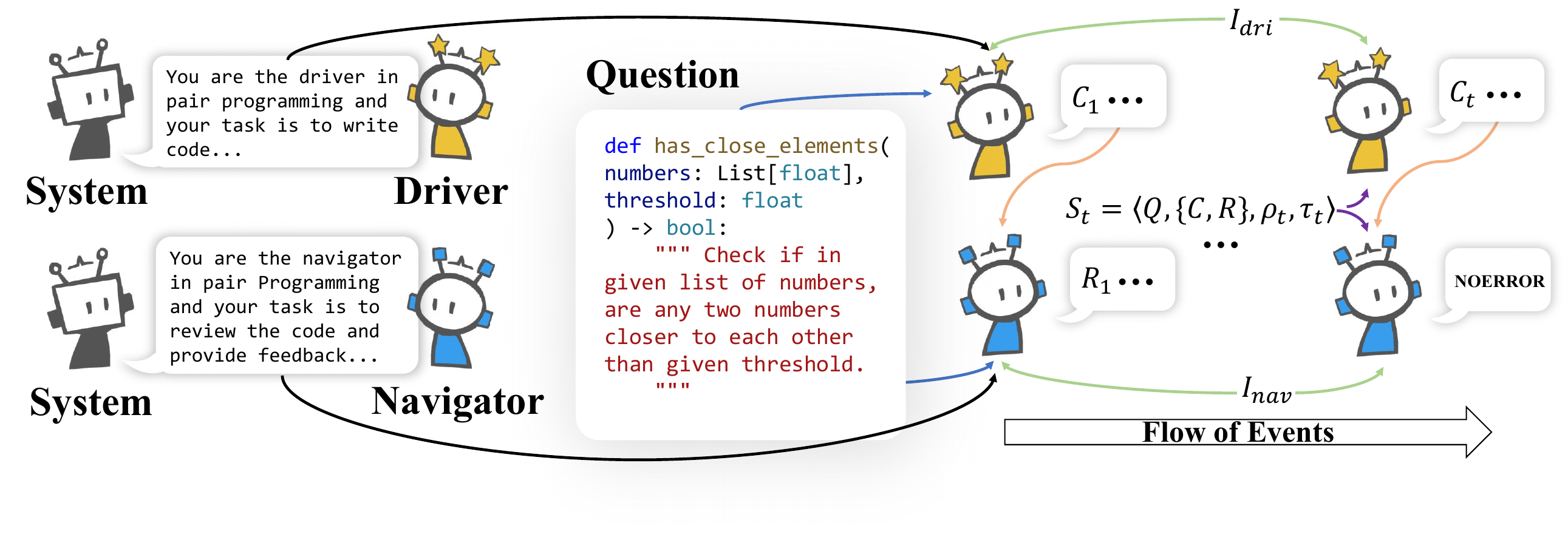}
    \vspace{-15mm}
    \caption{PairCoder workflow: the Driver proposes and revises, the Navigator reviews against toolchain evidence, and persistent errors trigger a role switch.}
    \label{events_flow}
\end{figure*}

\subsection{Role Assignment}\label{RoleAssignment}
At each time step, a role function $\rho: \mathcal{A} \times T \rightarrow \mathcal{R}$ assigns Driver and Navigator to the two agents, where $\mathcal{A} = \{a_1, a_2\}$ denotes the agents, $T$ the time steps, and $\mathcal{R} = \{Driver, Navigator\}$ the available roles.

The Driver is responsible for code construction and revision, whereas the Navigator checks correctness, identifies deficiencies, and returns actionable feedback. A \texttt{REVISE} verdict is itself an error signal, and the switch is taken \emph{before} the repair so that the agent that diagnosed the fault takes the keyboard and a different agent reviews the resulting fix. The default policy switches on each error ($\eta=1$); we also study switching only after $\eta$ consecutive errors, a fixed interval every $k$ rounds, and never switching (Sec.~\ref{sec:roleswitch}).

\paragraph{Driver Agent.}
The Driver generates code conditioned on the task specification and interaction history. As illustrated in Fig.~\ref{events_flow}, its generation process follows:
\begin{equation}\label{eq:1}
C_t = F_{driver}(I_{dri}, Q, M_{t-1})
\end{equation}
where $I_{dri}$ denotes the driver instructions, $Q$ is the task description, $C_t$ is the candidate code at iteration $t$, $R_i$ is the review outcome at iteration $i$, and $M_{t-1}$ is the accumulated interaction history:
\begin{equation}\label{eq:2}
M_t = \{(C_i, R_i) \mid 1 \le i \le t\}
\end{equation}

Concretely, on the first round the Driver is prompted with the task $Q$ alone and emits $C_1$; on every subsequent round it receives the previous candidate $C_{t-1}$, the Navigator's review $R_{t-1}$, and the same verification evidence $\psi_{t-1}$ the Navigator saw, and is asked to return the full revised program. There is no auxiliary objective or fine tuning: the Driver is a frozen LLM steered entirely through its prompt and the role-specific system message.

\paragraph{Navigator Agent}
The Navigator evaluates the current candidate against concrete verification evidence and returns structured feedback:
\begin{equation}\label{eq:3}
R_t = F_{navigator}(I_{nav}, Q, C_t, M_{t-1}, \psi_t)
\end{equation}
where $I_{nav}$ specifies navigator-specific review guidelines and $\psi_t$ is the verification evidence attached to $C_t$ (defined below).

The Navigator re-reads the requirements, traces the candidate on concrete inputs, checks that
names and interfaces match the specification, and weighs the evidence $\psi_t$, then issues one
of two verdicts:
\begin{equation}
R_t = \begin{cases}
\texttt{[NOERROR]} & \text{accept} \\
\text{REVISE}(\Delta_t) & \text{otherwise}
\end{cases}
\end{equation}
where $\Delta_t$ quotes the offending line, the requirement it violates, and the specific fix.
The protocol is deliberately conservative: when $\psi_t$ already passes and the Navigator cannot
point to a definite fault it must accept with \texttt{[NOERROR]} rather than churn working code
on style or vague suspicion. The interaction history is then updated as:
\begin{equation}\label{eq:4}
M_t = M_{t-1} \cup \{(C_t, R_t, \psi_t)\}
\end{equation}

\paragraph{Verification Evidence $\psi_t$.}
Before each review the candidate $C_t$ is run through the benchmark's own toolchain and the
result is attached to the Navigator's prompt as evidence $\psi_t$. In our implementation
$\psi_t$ combines up to three signals. (1) A compile or execute predicate from the renderer,
interpreter, or simulator (\LaTeX{}, CadQuery, matplotlib, the SVG rasterizer, Blender, or an
RTL simulator), together with the captured diagnostic on failure. (2) For tasks that ship a
specification, the outcome of a test the Navigator authors \emph{before} it reviews, a
test-driven check of $\mathrm{satisfy}(C_t,\mathrm{spec}(Q))$. (3) For image conditioned tasks,
a rendering of the current artifact placed beside the target, so the Navigator can cite
concrete visual differences. The Driver receives the same $\psi_t$ when it revises. This
grounding is what makes the second perspective informative rather than a paraphrase of the
first, and it is the only domain specific part of the loop; everything else is identical across
the seventeen benchmarks.

\subsection{System Components}

\paragraph{Shared Environment.}
The framework maintains a shared state $S_t = \langle Q, M_t, \rho_t, \tau_t \rangle$ with four components. The query $Q$ is the original task specification. The memory $M_t$ is the running interaction history (candidates, reviews, and the attached evidence), kept per role and bounded in practice only by the model's context window. The role assignment $\rho_t$ records the current mapping between agents and roles, and the iteration counter $\tau_t$ tracks collaboration progress.

\paragraph{Prompt construction.}
Each agent is a frozen chat LLM accessed through an OpenAI compatible API and carries its own
message history; the two histories together realize the shared memory $M_t$. The role prompt is
injected as a system message, a one line Driver or Navigator instruction plus an optional
benchmark specific hint, and is re-asserted whenever the roles switch, which is the Self-Mirror
mechanism in implementation terms. The Driver's user message is the task $Q$ on the first round
and the previous program, the Navigator's review, and the evidence $\psi_{t-1}$ on every round
after; the Navigator's user message is a review instruction, the candidate program, and
$\psi_t$. For image conditioned benchmarks these messages are multimodal: the Driver is shown
the target image, and the Navigator is additionally shown a rendering of the current candidate
next to the target. The single model baseline is one direct generation with the same task prompt
and no review.

\begin{figure}[htbp]
    \centering
    \includegraphics[width=0.44\textwidth]{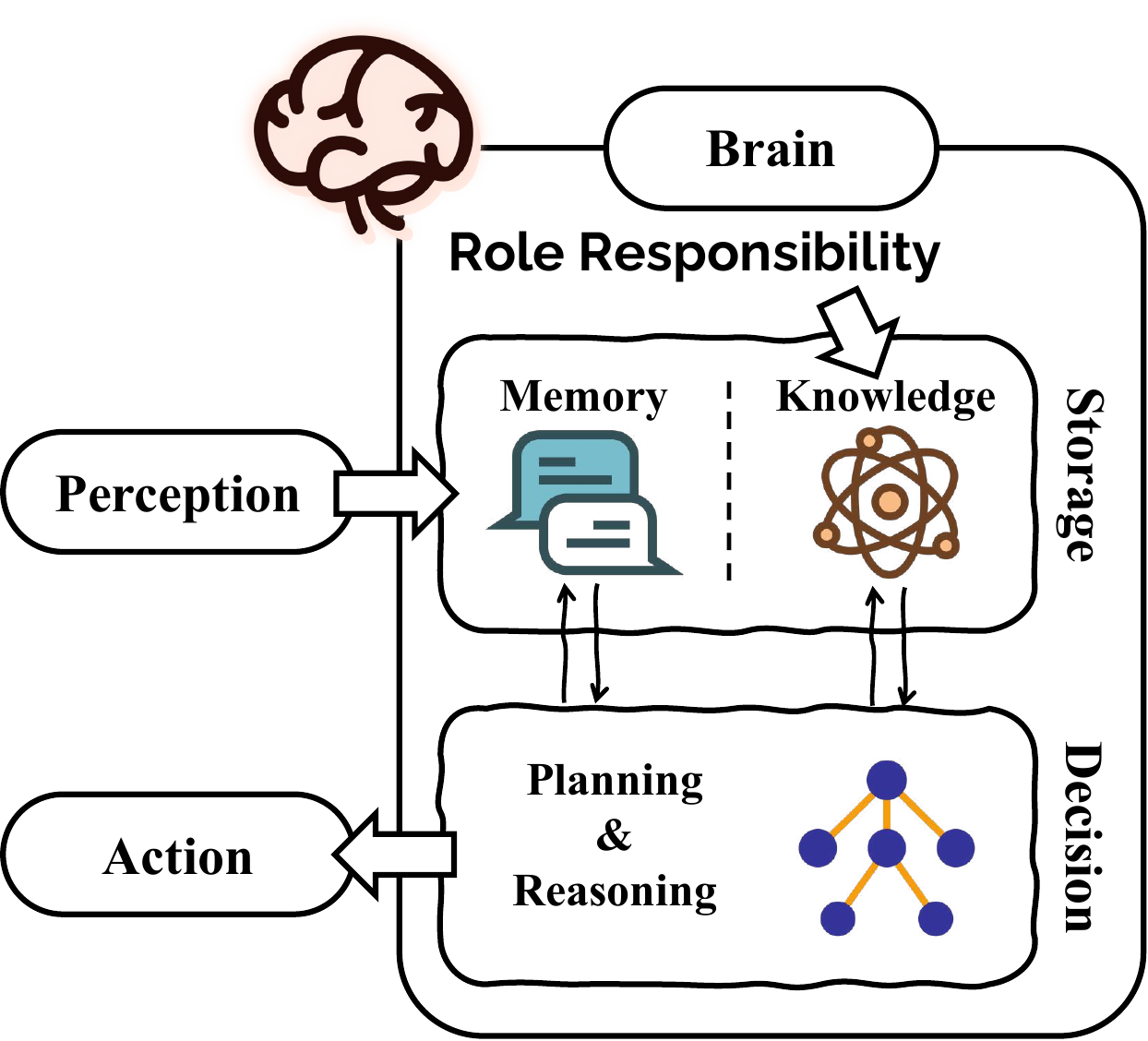}
    \caption{Agent brain architecture: LLM-based decision making with role-specific configuration.}
    \label{agent_brain}
\end{figure}

\paragraph{Agent Brain.}
As depicted in Fig.~\ref{agent_brain}, each agent uses an LLM conditioned on a role specific prompt and the current shared state:
\begin{equation}
\text{Brain}_{\text{role}} = \text{LLM}(P_{\text{role}}, S_t)
\end{equation}
where $P_{\text{role}}$ encodes the responsibilities of the current role and $S_t$ provides the task context, dialogue history, and current role assignment.

\paragraph{Action Space.}
Each agent operates within a discrete action space $\mathcal{A}_{\text{role}}$:
\begin{align}
\mathcal{A}_{driver} &= \{\textsc{Generate},\ \textsc{Refine},\ \textsc{End}\} \\
\mathcal{A}_{navigator} &= \{\textsc{Review},\ \textsc{Accept}\}
\end{align}

\subsection{Collaboration Protocol}
\begin{figure}[t]
    \centering
    \includegraphics[width=0.45\textwidth]{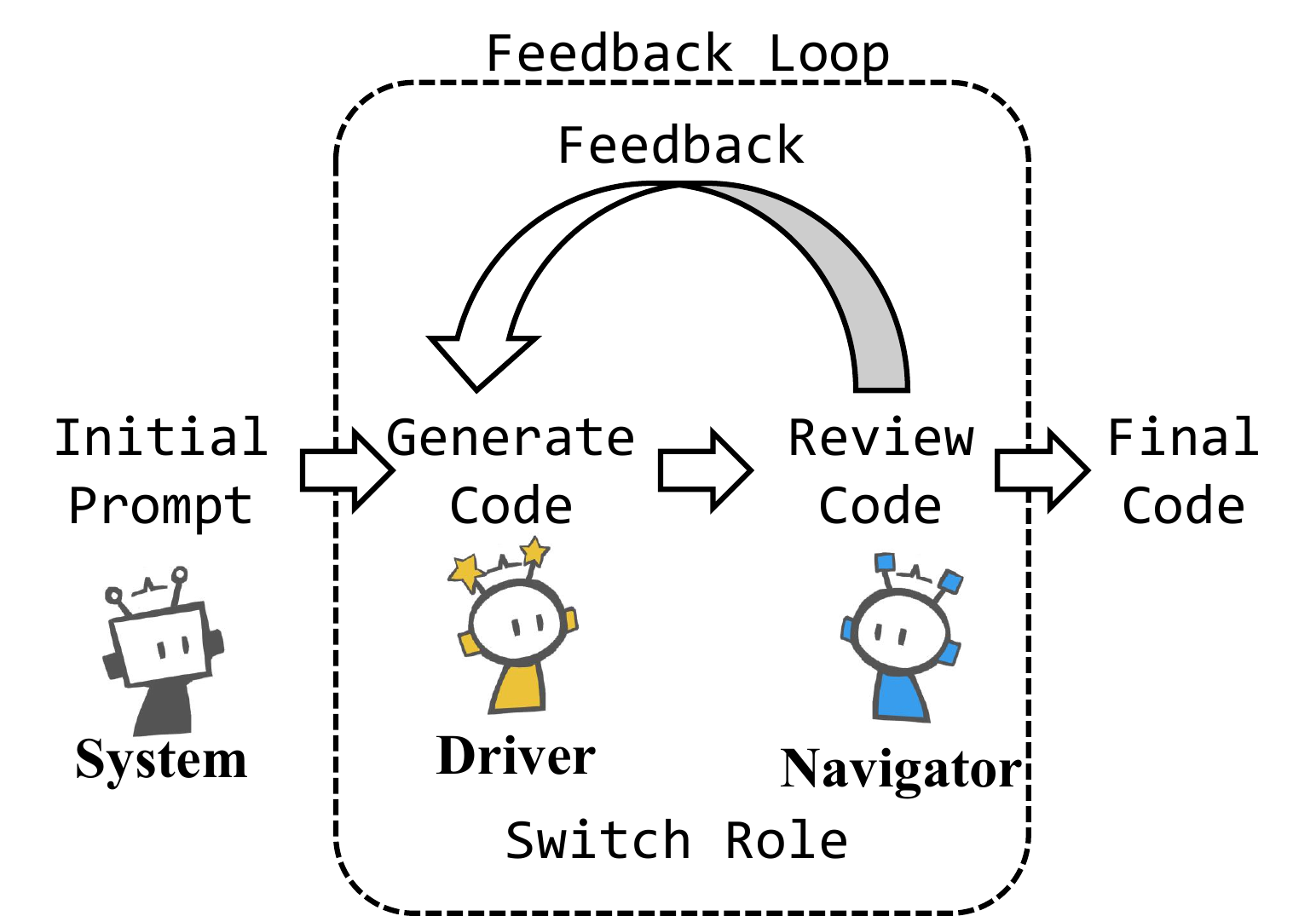}
    \caption{Collaboration mechanism with iterative feedback loop between Driver and Navigator agents.}
    \label{coll_mech}
\end{figure}

The collaboration follows an iterative refinement process formalized in Algorithm~\ref{alg:paircoder}. The feedback loop, illustrated in Fig.~\ref{coll_mech}, alternates between Navigator review and Driver revision, yielding progressive improvement over successive iterations.

\begin{algorithm}[t]
\caption{PairCoder Collaboration Protocol}
\label{alg:paircoder}
\begin{algorithmic}[1]
\State \textbf{Input:} Task $Q$, parameters $\Theta = \{T, k, \eta\}$ (budget, fixed interval, error threshold)
\State \textbf{Output:} Optimized code $C^*$
\State Initialize $S_0 \leftarrow \langle Q, \emptyset, \rho_0, 0 \rangle$
\While{$\tau < T$ and not $\text{terminated}$}
    \State \textit{Driver phase with self mirror}
    \State $P_{dri} \leftarrow \text{SelfMirror}(\text{Driver}, I_{dri})$
    \State $C_\tau \leftarrow F_{driver}(P_{dri}, Q, M_{\tau-1})$
    \State \textit{Navigator phase with self mirror}
    \State $\psi_\tau \leftarrow \text{Verify}(C_\tau)$ \Comment{compile/execute, opt.\ test, opt.\ render}
    \State $P_{nav} \leftarrow \text{SelfMirror}(\text{Navigator}, I_{nav})$
    \State $R_\tau \leftarrow F_{navigator}(P_{nav}, Q, C_\tau, M_{\tau-1}, \psi_\tau)$
    \If{$R_\tau = \texttt{[NOERROR]}$}
        \State \Return $C_\tau$
    \EndIf
    \State \textit{Update shared state}
    \State $M_\tau \leftarrow \text{Update}(M_{\tau-1}, C_\tau, R_\tau)$
    \State $\rho_{\tau+1} \leftarrow \text{CheckSwitch}(\rho_\tau, R_\tau, \tau)$
    \State $\tau \leftarrow \tau + 1$
\EndWhile
\State \Return $\arg\max_{C_i \in M_\tau} \text{Quality}(C_i)$
\end{algorithmic}
\end{algorithm}

\paragraph{Termination Protocol.}
The system terminates when condition $\Omega(S_t)$ is satisfied:
\begin{equation}
\Omega(S_t) = \begin{cases}
\text{true} & \text{if } R_t = \texttt{[NOERROR]} \\
\text{true} & \text{if } \tau \geq T \\
\text{false} & \text{otherwise}
\end{cases}
\end{equation}
If the loop stops on \texttt{[NOERROR]} the accepted candidate is returned. If instead the
budget $T$ is exhausted, PairCoder does not return the last attempt but the best one seen,
$\arg\max_{C_i \in M_\tau}\mathrm{Quality}(C_i)$, where $\mathrm{Quality}$ orders candidates
first by whether the verification evidence $\psi$ passes, then by the benchmark's continuous
score when one is available (for example structural similarity to the target), and finally by
recency. This makes an unresolved review debate degrade to a safe fallback rather than a
failure.

\paragraph{Self-Mirror Mechanism.}
To preserve role consistency, each agent receives an identity aware prompt before generation or review:
\begin{equation}
\begin{aligned}
\text{SelfMirror}(\rho_t, Q, M_t)
&= \text{concat}( \\
&\quad \text{RolePrompt}(\rho_t), Q, M_t)
\end{aligned}
\end{equation}
where $\text{RolePrompt}(\rho_t)$ explicitly states whether the agent should act as Driver or Navigator. This mechanism keeps the Driver focused on code production and revision, and keeps the Navigator focused on diagnosis, verification, and targeted feedback throughout the interaction.

\paragraph{Implementation.}
Both arms use identical prompts and the same per benchmark extractor, which recovers the program
from the raw reply only to run the verifier $\psi$; the dialogue itself keeps the raw text.
Empty or failed API responses are retried with exponential backoff and do not pollute the
history. Reasoning is disabled for the main grid (a thinking-enabled study is reported
separately), and the loop uses at most three to four review rounds in the main experiments (up
to eight in the role-switching study of Sec.~\ref{sec:roleswitch}). The verifier $\psi$ is the
only per benchmark component; it wraps the benchmark's official toolchain (compiler, interpreter,
renderer, simulator, or unit tests). We log prompt and completion tokens separately for each
arm, which is the basis for the token accounting in Sec.~\ref{sec:tokens}.

\paragraph{Role Switching Strategies.}
Our framework supports heterogeneous agent configurations in which different LLMs can serve as Driver and Navigator, allowing us to leverage complementary model strengths. To optimize collaboration dynamics in this setting, we implement two switching policies $\pi: S_t \rightarrow \{0, 1\}$:

\textbf{Fixed interval switching.}
\begin{equation}
\pi_{\text{fixed}}(S_t) = (\tau \bmod k = 0)
\end{equation}
ensuring balanced participation every $k$ iterations.

\textbf{Error triggered switching.}
\begin{equation}
\pi_{\text{error}}(S_t) = \left(\sum_{i=\tau-w}^{\tau} \mathbb{I}[R_i = \text{REVISE}] \geq \eta\right)
\end{equation}
activating when error count exceeds threshold $\eta$ within window $w$.

\section{Experiments}
\label{sec:experiments}

\subsection{Experimental Setup}
\label{sec:setup}
\paragraph{Benchmarks.}
Systematic LLM evaluation has been studied across many domains, from Chinese language
understanding~\cite{zhujiu} and strategic reasoning~\cite{llmspark} to web interface
comprehension~\cite{iwbench}. Following this tradition, we evaluate PairCoder on 17 public
benchmarks organized by the artifact a program must produce. \textit{Program synthesis}:
LiveCodeBench~\cite{jain2024livecodebench} (recent contest problems),
BigCodeBench~\cite{zhuo2024bigcodebench} (library intensive tasks), and
DS-1000~\cite{lai2023ds1000} (data science). \textit{Multilingual code}: the C++, Java, and
JavaScript splits of HumanEval-X~\cite{zheng2023codegeex}. \textit{Web applications}:
WebApp1K~\cite{cui2024webapp1k} (React user journeys scored by hidden Jest tests).
\textit{Hardware}: VerilogEval~\cite{liu2023verilogeval} and RTLLM~\cite{lu2024rtllm}
(simulation against reference testbenches). \textit{Code driven artifacts}:
DaTikZ~\cite{belouadi2024automatikz} (caption to Ti\textit{k}Z),
Plot2Code~\cite{wu2024plot2code}, PandasPlotBench~\cite{galimzyanov2024drawing}, and
ChartMimic~\cite{yang2024chartmimic} (chart or data to matplotlib),
StarVector~\cite{rodriguez2024starvector} (image to SVG), GenCAD-Code~\cite{alam2024gencad}
(image to CadQuery), 3DCodeBench~\cite{gao20263dcodebench} (text to Blender script), and
P3D-Bench~\cite{p3dbench2026} (text to parametric 3D CAD, detailed in Sec.~\ref{sec:p3dbench}). Each
benchmark is scored with its official metric suite: pass@1 or execution rate where unit tests
or toolchains define correctness, plus SSIM~\cite{wang2004image},
CLIP~\cite{radford2021learning}, DINO~\cite{oquab2023dinov2},
SigLIP-2~\cite{tschannen2025siglip}, and Chamfer distance for rendered artifacts; full per benchmark scoring and reproducibility details are in Appendix~\ref{app:repro}.

\paragraph{Models and protocol.}
We test seven models from three vendors: \texttt{gpt-5.4-mini}, \texttt{gpt-5.4}, and
\texttt{gpt-5.5} (OpenAI); \texttt{doubao-1.5-lite} and \texttt{doubao-seed-2.0-mini}
(Doubao); \texttt{deepseek-v3.2} and \texttt{deepseek-v4-flash} (DeepSeek), all with thinking
disabled. The baseline is a single direct generation (one call per problem). PairCoder runs
the loop of Sec.~\ref{RoleAssignment} with at most three to four review rounds: the Driver
generates, the verification predicate $\psi$ compiles, executes, or renders the candidate, the
Navigator reviews the code together with this evidence (and, for visually grounded tasks, a
rendering of the current artifact next to the target) and either accepts with
\texttt{[NOERROR]} or requests a concrete revision, and roles switch when errors persist. Both
arms use identical prompts and extraction; image input benchmarks are restricted to the
vision capable models. We log every token in both arms.

\providecolor{improvebg}{RGB}{226,242,228}
\begin{table*}[htbp]
\caption{Main results: 17 benchmarks $\times$ seven models from three vendors (thinking off),
each scored with its official metric suite. Each cell is single model $\rightarrow$
\textbf{PairCoder}; bold marks the better arm ($\downarrow$ metrics are better when lower;
visual metrics are aggregates that score non-rendering generations as 0), green marks
improvements. $\dagger$ marks image-input benchmarks under vision-less models
(\texttt{doubao-1.5-lite}, \texttt{deepseek-v3.2}, \texttt{deepseek-v4-flash}). \textbf{P3D-Bench} denotes its \emph{text-to-3D / minimal-JSON} track
(Sec.~\ref{sec:p3dbench}); image-to-3D and assembly-3D are reported as a demonstration in
Table~\ref{tab:p3d_demo}.}
\label{tab:main_grid}
\centering
\renewcommand{\arraystretch}{1.13}
\setlength{\tabcolsep}{3.2pt}
\newcolumntype{C}{>{\centering\arraybackslash}p{2.45cm}}
\resizebox{0.995\textwidth}{!}{%
\begin{tabular}{ll*{7}{C}}
\toprule
\textbf{Benchmark} & \textbf{Metric} & \shortstack[c]{\texttt{gpt-5.4-}\\\texttt{mini}} & \shortstack[c]{\texttt{gpt-5.4}\\\strut} & \shortstack[c]{\texttt{gpt-5.5}\\\strut} & \shortstack[c]{\texttt{doubao-}\\\texttt{1.5-lite}} & \shortstack[c]{\texttt{doubao-seed-}\\\texttt{2.0-mini}} & \shortstack[c]{\texttt{deepseek-}\\\texttt{v3.2}} & \shortstack[c]{\texttt{deepseek-}\\\texttt{v4-flash}} \\
\midrule
\multicolumn{9}{l}{\textit{\textbf{Program synthesis}}} \\
\cmidrule(lr){1-9}
LiveCodeBench & pass@1 $\uparrow$ & \cellcolor{improvebg}0.942$\rightarrow$\textbf{0.992} & \cellcolor{improvebg}0.917$\rightarrow$\textbf{0.958} & \cellcolor{improvebg}0.933$\rightarrow$\textbf{0.975} & \cellcolor{improvebg}0.500$\rightarrow$\textbf{0.533} & \cellcolor{improvebg}0.825$\rightarrow$\textbf{0.867} & 0.950$\rightarrow$0.908 & \cellcolor{improvebg}0.933$\rightarrow$\textbf{0.975} \\
\cmidrule(lr){1-9}
BigCodeBench & pass@1 $\uparrow$ & \cellcolor{improvebg}0.450$\rightarrow$\textbf{0.458} & \cellcolor{improvebg}0.433$\rightarrow$\textbf{0.450} & 0.458$\rightarrow$0.458 & 0.350$\rightarrow$0.317 & 0.408$\rightarrow$0.392 & 0.467$\rightarrow$0.450 & \cellcolor{improvebg}0.417$\rightarrow$\textbf{0.425} \\
\cmidrule(lr){1-9}
DS-1000 & pass@1 $\uparrow$ & \cellcolor{improvebg}0.250$\rightarrow$\textbf{0.255} & \cellcolor{improvebg}0.285$\rightarrow$\textbf{0.310} & \cellcolor{improvebg}0.385$\rightarrow$\textbf{0.425} & 0.075$\rightarrow$0.075 & 0.100$\rightarrow$0.085 & 0.155$\rightarrow$0.080 & 0.170$\rightarrow$0.110 \\
\midrule
\multicolumn{9}{l}{\textit{\textbf{Multilingual}}} \\
\cmidrule(lr){1-9}
HumanEval-X C++ & pass@1 $\uparrow$ & \cellcolor{improvebg}0.335$\rightarrow$\textbf{0.415} & \cellcolor{improvebg}0.463$\rightarrow$\textbf{0.573} & \cellcolor{improvebg}0.524$\rightarrow$\textbf{0.835} & \cellcolor{improvebg}0.183$\rightarrow$\textbf{0.220} & 0.427$\rightarrow$0.421 & 0.463$\rightarrow$0.451 & \cellcolor{improvebg}0.561$\rightarrow$\textbf{0.598} \\
\cmidrule(lr){1-9}
HumanEval-X Java & pass@1 $\uparrow$ & \cellcolor{improvebg}0.341$\rightarrow$\textbf{0.384} & \cellcolor{improvebg}0.451$\rightarrow$\textbf{0.500} & \cellcolor{improvebg}0.396$\rightarrow$\textbf{0.561} & \cellcolor{improvebg}0.250$\rightarrow$\textbf{0.256} & 0.335$\rightarrow$0.335 & 0.500$\rightarrow$0.463 & 0.494$\rightarrow$0.463 \\
\cmidrule(lr){1-9}
HumanEval-X JS & pass@1 $\uparrow$ & \cellcolor{improvebg}0.323$\rightarrow$\textbf{0.341} & \cellcolor{improvebg}0.445$\rightarrow$\textbf{0.500} & \cellcolor{improvebg}0.384$\rightarrow$\textbf{0.543} & 0.256$\rightarrow$0.250 & \cellcolor{improvebg}0.293$\rightarrow$\textbf{0.305} & \cellcolor{improvebg}0.463$\rightarrow$\textbf{0.482} & \cellcolor{improvebg}0.317$\rightarrow$\textbf{0.366} \\
\midrule
\multicolumn{9}{l}{\textit{\textbf{Web \& Hardware}}} \\
\cmidrule(lr){1-9}
WebApp1K & pass@1 $\uparrow$ & \cellcolor{improvebg}0.700$\rightarrow$\textbf{0.838} & \cellcolor{improvebg}0.787$\rightarrow$\textbf{0.838} & \cellcolor{improvebg}0.787$\rightarrow$\textbf{0.887} & 0.013$\rightarrow$0.013 & \cellcolor{improvebg}0.525$\rightarrow$\textbf{0.750} & \cellcolor{improvebg}0.575$\rightarrow$\textbf{0.750} & \cellcolor{improvebg}0.650$\rightarrow$\textbf{0.675} \\
\cmidrule(lr){1-9}
VerilogEval & pass@1 $\uparrow$ & \cellcolor{improvebg}0.763$\rightarrow$\textbf{0.808} & \cellcolor{improvebg}0.737$\rightarrow$\textbf{0.827} & \cellcolor{improvebg}0.821$\rightarrow$\textbf{0.897} & \cellcolor{improvebg}0.429$\rightarrow$\textbf{0.436} & \cellcolor{improvebg}0.647$\rightarrow$\textbf{0.699} & \cellcolor{improvebg}0.596$\rightarrow$\textbf{0.724} & \cellcolor{improvebg}0.692$\rightarrow$\textbf{0.705} \\
\cmidrule(lr){1-9}
RTLLM & pass@1 $\uparrow$ & \cellcolor{improvebg}0.656$\rightarrow$\textbf{0.719} & \cellcolor{improvebg}0.400$\rightarrow$\textbf{0.620} & \cellcolor{improvebg}0.620$\rightarrow$\textbf{0.660} & 0.360$\rightarrow$0.300 & 0.500$\rightarrow$0.480 & \cellcolor{improvebg}0.560$\rightarrow$\textbf{0.640} & 0.560$\rightarrow$0.520 \\
\midrule
\multicolumn{9}{l}{\textit{\textbf{Code-driven artifacts}}} \\
\cmidrule(lr){1-9}
\multirow{4}{*}{DaTikZ} & compile $\uparrow$ & \cellcolor{improvebg}0.500$\rightarrow$\textbf{0.633} & \cellcolor{improvebg}0.717$\rightarrow$\textbf{0.967} & \cellcolor{improvebg}0.783$\rightarrow$\textbf{1.000} & \cellcolor{improvebg}0.550$\rightarrow$\textbf{0.850} & \cellcolor{improvebg}0.567$\rightarrow$\textbf{0.800} & \cellcolor{improvebg}0.683$\rightarrow$\textbf{0.917} & \cellcolor{improvebg}0.633$\rightarrow$\textbf{0.733} \\
 & SSIM $\uparrow$ & \cellcolor{improvebg}0.215$\rightarrow$\textbf{0.256} & \cellcolor{improvebg}0.363$\rightarrow$\textbf{0.512} & \cellcolor{improvebg}0.394$\rightarrow$\textbf{0.485} & \cellcolor{improvebg}0.406$\rightarrow$\textbf{0.601} & \cellcolor{improvebg}0.325$\rightarrow$\textbf{0.449} & \cellcolor{improvebg}0.365$\rightarrow$\textbf{0.544} & \cellcolor{improvebg}0.348$\rightarrow$\textbf{0.406} \\
 & CLIP $\uparrow$ & \cellcolor{improvebg}0.327$\rightarrow$\textbf{0.384} & \cellcolor{improvebg}0.566$\rightarrow$\textbf{0.770} & \cellcolor{improvebg}0.602$\rightarrow$\textbf{0.767} & \cellcolor{improvebg}0.391$\rightarrow$\textbf{0.621} & \cellcolor{improvebg}0.441$\rightarrow$\textbf{0.606} & \cellcolor{improvebg}0.508$\rightarrow$\textbf{0.702} & \cellcolor{improvebg}0.505$\rightarrow$\textbf{0.578} \\
 & DINO $\uparrow$ & \cellcolor{improvebg}0.303$\rightarrow$\textbf{0.338} & \cellcolor{improvebg}0.530$\rightarrow$\textbf{0.698} & \cellcolor{improvebg}0.554$\rightarrow$\textbf{0.730} & \cellcolor{improvebg}0.259$\rightarrow$\textbf{0.391} & \cellcolor{improvebg}0.383$\rightarrow$\textbf{0.545} & \cellcolor{improvebg}0.465$\rightarrow$\textbf{0.596} & \cellcolor{improvebg}0.445$\rightarrow$\textbf{0.519} \\
\cmidrule(lr){1-9}
\multirow{3}{*}{Plot2Code} & exec $\uparrow$ & \cellcolor{improvebg}0.841$\rightarrow$\textbf{0.962} & \cellcolor{improvebg}0.962$\rightarrow$\textbf{0.977} & 0.985$\rightarrow$0.977 & $\dagger$ & \cellcolor{improvebg}0.909$\rightarrow$\textbf{0.917} & $\dagger$ & $\dagger$ \\
 & SSIM $\uparrow$ & \cellcolor{improvebg}0.412$\rightarrow$\textbf{0.474} & \cellcolor{improvebg}0.476$\rightarrow$\textbf{0.495} & \cellcolor{improvebg}0.487$\rightarrow$\textbf{0.496} & $\dagger$ & \cellcolor{improvebg}0.410$\rightarrow$\textbf{0.443} & $\dagger$ & $\dagger$ \\
 & CLIP $\uparrow$ & \cellcolor{improvebg}0.802$\rightarrow$\textbf{0.912} & \cellcolor{improvebg}0.916$\rightarrow$\textbf{0.920} & 0.943$\rightarrow$0.931 & $\dagger$ & 0.835$\rightarrow$0.823 & $\dagger$ & $\dagger$ \\
\cmidrule(lr){1-9}
\multirow{3}{*}{PandasPlotBench} & exec $\uparrow$ & \cellcolor{improvebg}0.789$\rightarrow$\textbf{0.851} & \cellcolor{improvebg}0.966$\rightarrow$\textbf{0.994} & \cellcolor{improvebg}0.960$\rightarrow$\textbf{0.977} & $\dagger$ & \cellcolor{improvebg}0.909$\rightarrow$\textbf{0.989} & $\dagger$ & $\dagger$ \\
 & SSIM $\uparrow$ & \cellcolor{improvebg}0.423$\rightarrow$\textbf{0.461} & \cellcolor{improvebg}0.567$\rightarrow$\textbf{0.610} & \cellcolor{improvebg}0.597$\rightarrow$\textbf{0.644} & $\dagger$ & \cellcolor{improvebg}0.491$\rightarrow$\textbf{0.534} & $\dagger$ & $\dagger$ \\
 & CLIP $\uparrow$ & \cellcolor{improvebg}0.728$\rightarrow$\textbf{0.794} & \cellcolor{improvebg}0.909$\rightarrow$\textbf{0.944} & \cellcolor{improvebg}0.907$\rightarrow$\textbf{0.942} & $\dagger$ & \cellcolor{improvebg}0.846$\rightarrow$\textbf{0.922} & $\dagger$ & $\dagger$ \\
\cmidrule(lr){1-9}
\multirow{3}{*}{ChartMimic} & exec $\uparrow$ & 0.967$\rightarrow$0.967 & \cellcolor{improvebg}0.917$\rightarrow$\textbf{0.983} & \cellcolor{improvebg}0.983$\rightarrow$\textbf{1.000} & $\dagger$ & \cellcolor{improvebg}0.900$\rightarrow$\textbf{0.950} & $\dagger$ & $\dagger$ \\
 & SSIM $\uparrow$ & \cellcolor{improvebg}0.578$\rightarrow$\textbf{0.578} & \cellcolor{improvebg}0.533$\rightarrow$\textbf{0.582} & \cellcolor{improvebg}0.593$\rightarrow$\textbf{0.613} & $\dagger$ & \cellcolor{improvebg}0.443$\rightarrow$\textbf{0.476} & $\dagger$ & $\dagger$ \\
 & CLIP $\uparrow$ & 0.871$\rightarrow$0.861 & \cellcolor{improvebg}0.820$\rightarrow$\textbf{0.848} & \cellcolor{improvebg}0.874$\rightarrow$\textbf{0.887} & $\dagger$ & \cellcolor{improvebg}0.770$\rightarrow$\textbf{0.808} & $\dagger$ & $\dagger$ \\
\cmidrule(lr){1-9}
\multirow{4}{*}{StarVector} & render $\uparrow$ & \cellcolor{improvebg}0.983$\rightarrow$\textbf{1.000} & \cellcolor{improvebg}0.900$\rightarrow$\textbf{1.000} & 1.000$\rightarrow$1.000 & $\dagger$ & \cellcolor{improvebg}0.983$\rightarrow$\textbf{1.000} & $\dagger$ & $\dagger$ \\
 & SSIM $\uparrow$ & \cellcolor{improvebg}0.784$\rightarrow$\textbf{0.801} & \cellcolor{improvebg}0.776$\rightarrow$\textbf{0.873} & \cellcolor{improvebg}0.777$\rightarrow$\textbf{0.787} & $\dagger$ & \cellcolor{improvebg}0.758$\rightarrow$\textbf{0.770} & $\dagger$ & $\dagger$ \\
 & CLIP $\uparrow$ & \cellcolor{improvebg}0.929$\rightarrow$\textbf{0.944} & \cellcolor{improvebg}0.859$\rightarrow$\textbf{0.954} & 0.961$\rightarrow$0.960 & $\dagger$ & \cellcolor{improvebg}0.883$\rightarrow$\textbf{0.915} & $\dagger$ & $\dagger$ \\
 & DINO $\uparrow$ & \cellcolor{improvebg}0.874$\rightarrow$\textbf{0.885} & \cellcolor{improvebg}0.820$\rightarrow$\textbf{0.906} & 0.938$\rightarrow$0.934 & $\dagger$ & \cellcolor{improvebg}0.793$\rightarrow$\textbf{0.819} & $\dagger$ & $\dagger$ \\
\cmidrule(lr){1-9}
\multirow{2}{*}{GenCAD-Code} & exec $\uparrow$ & \cellcolor{improvebg}0.900$\rightarrow$\textbf{0.983} & \cellcolor{improvebg}0.867$\rightarrow$\textbf{0.983} & \cellcolor{improvebg}0.917$\rightarrow$\textbf{1.000} & $\dagger$ & \cellcolor{improvebg}0.833$\rightarrow$\textbf{0.967} & $\dagger$ & $\dagger$ \\
 & Chamfer $\downarrow$ & \cellcolor{improvebg}0.233$\rightarrow$\textbf{0.166} & \cellcolor{improvebg}0.259$\rightarrow$\textbf{0.155} & \cellcolor{improvebg}0.215$\rightarrow$\textbf{0.126} & $\dagger$ & \cellcolor{improvebg}0.308$\rightarrow$\textbf{0.212} & $\dagger$ & $\dagger$ \\
\cmidrule(lr){1-9}
\multirow{4}{*}{3DCodeBench} & exec $\uparrow$ & \cellcolor{improvebg}0.200$\rightarrow$\textbf{0.783} & \cellcolor{improvebg}0.433$\rightarrow$\textbf{0.783} & \cellcolor{improvebg}0.383$\rightarrow$\textbf{0.417} & \cellcolor{improvebg}0.167$\rightarrow$\textbf{0.383} & \cellcolor{improvebg}0.200$\rightarrow$\textbf{0.533} & \cellcolor{improvebg}0.333$\rightarrow$\textbf{0.633} & 0.600$\rightarrow$0.533 \\
 & SigLIP-2 $\uparrow$ & \cellcolor{improvebg}0.181$\rightarrow$\textbf{0.702} & \cellcolor{improvebg}0.263$\rightarrow$\textbf{0.842} & \cellcolor{improvebg}0.741$\rightarrow$\textbf{0.790} & \cellcolor{improvebg}0.225$\rightarrow$\textbf{0.302} & \cellcolor{improvebg}0.180$\rightarrow$\textbf{0.521} & \cellcolor{improvebg}0.271$\rightarrow$\textbf{0.446} & 0.658$\rightarrow$0.487 \\
 & DINO $\uparrow$ & \cellcolor{improvebg}0.101$\rightarrow$\textbf{0.376} & \cellcolor{improvebg}0.129$\rightarrow$\textbf{0.512} & \cellcolor{improvebg}0.362$\rightarrow$\textbf{0.420} & 0.138$\rightarrow$0.132 & \cellcolor{improvebg}0.097$\rightarrow$\textbf{0.300} & \cellcolor{improvebg}0.175$\rightarrow$\textbf{0.261} & 0.372$\rightarrow$0.290 \\
 & Chamfer $\downarrow$ & \cellcolor{improvebg}4.850$\rightarrow$\textbf{2.618} & \cellcolor{improvebg}4.658$\rightarrow$\textbf{1.658} & 2.368$\rightarrow$2.482 & \cellcolor{improvebg}4.762$\rightarrow$\textbf{3.955} & \cellcolor{improvebg}4.965$\rightarrow$\textbf{3.400} & \cellcolor{improvebg}5.279$\rightarrow$\textbf{3.407} & 2.694$\rightarrow$3.258 \\
\cmidrule(lr){1-9}
\multirow{7}{*}{\makecell[l]{P3D-Bench\\\scriptsize text-to-3D\\\scriptsize (minimal-JSON)}} & valid $\uparrow$ & \cellcolor{improvebg}0.973$\rightarrow$\textbf{1.000} & \cellcolor{improvebg}0.715$\rightarrow$\textbf{0.990} & \cellcolor{improvebg}0.672$\rightarrow$\textbf{0.943} & \cellcolor{improvebg}0.140$\rightarrow$\textbf{0.427} & \cellcolor{improvebg}0.812$\rightarrow$\textbf{0.990} & \cellcolor{improvebg}0.950$\rightarrow$\textbf{0.993} & \cellcolor{improvebg}0.945$\rightarrow$\textbf{1.000} \\
 & geometry $\uparrow$ & \cellcolor{improvebg}0.600$\rightarrow$\textbf{0.618} & \cellcolor{improvebg}0.347$\rightarrow$\textbf{0.483} & \cellcolor{improvebg}0.321$\rightarrow$\textbf{0.438} & \cellcolor{improvebg}0.073$\rightarrow$\textbf{0.218} & \cellcolor{improvebg}0.419$\rightarrow$\textbf{0.497} & \cellcolor{improvebg}0.549$\rightarrow$\textbf{0.576} & \cellcolor{improvebg}0.581$\rightarrow$\textbf{0.611} \\
 & F@0.05 $\uparrow$ & \cellcolor{improvebg}0.746$\rightarrow$\textbf{0.767} & \cellcolor{improvebg}0.443$\rightarrow$\textbf{0.619} & \cellcolor{improvebg}0.419$\rightarrow$\textbf{0.570} & \cellcolor{improvebg}0.094$\rightarrow$\textbf{0.275} & \cellcolor{improvebg}0.537$\rightarrow$\textbf{0.633} & \cellcolor{improvebg}0.682$\rightarrow$\textbf{0.714} & \cellcolor{improvebg}0.727$\rightarrow$\textbf{0.764} \\
 & F@0.01 $\uparrow$ & \cellcolor{improvebg}0.316$\rightarrow$\textbf{0.326} & \cellcolor{improvebg}0.153$\rightarrow$\textbf{0.215} & \cellcolor{improvebg}0.143$\rightarrow$\textbf{0.188} & \cellcolor{improvebg}0.037$\rightarrow$\textbf{0.104} & \cellcolor{improvebg}0.196$\rightarrow$\textbf{0.230} & \cellcolor{improvebg}0.282$\rightarrow$\textbf{0.298} & \cellcolor{improvebg}0.298$\rightarrow$\textbf{0.311} \\
 & NC $\uparrow$ & \cellcolor{improvebg}0.678$\rightarrow$\textbf{0.698} & \cellcolor{improvebg}0.435$\rightarrow$\textbf{0.605} & \cellcolor{improvebg}0.402$\rightarrow$\textbf{0.556} & \cellcolor{improvebg}0.087$\rightarrow$\textbf{0.262} & \cellcolor{improvebg}0.504$\rightarrow$\textbf{0.609} & \cellcolor{improvebg}0.640$\rightarrow$\textbf{0.671} & \cellcolor{improvebg}0.650$\rightarrow$\textbf{0.685} \\
 & IoU $\uparrow$ & \cellcolor{improvebg}0.486$\rightarrow$\textbf{0.501} & \cellcolor{improvebg}0.226$\rightarrow$\textbf{0.334} & \cellcolor{improvebg}0.207$\rightarrow$\textbf{0.288} & \cellcolor{improvebg}0.054$\rightarrow$\textbf{0.157} & \cellcolor{improvebg}0.302$\rightarrow$\textbf{0.360} & \cellcolor{improvebg}0.434$\rightarrow$\textbf{0.460} & \cellcolor{improvebg}0.474$\rightarrow$\textbf{0.498} \\
 & topology $\uparrow$ & \cellcolor{improvebg}0.968$\rightarrow$\textbf{0.995} & \cellcolor{improvebg}0.708$\rightarrow$\textbf{0.983} & \cellcolor{improvebg}0.669$\rightarrow$\textbf{0.938} & \cellcolor{improvebg}0.136$\rightarrow$\textbf{0.416} & \cellcolor{improvebg}0.808$\rightarrow$\textbf{0.985} & \cellcolor{improvebg}0.946$\rightarrow$\textbf{0.989} & \cellcolor{improvebg}0.943$\rightarrow$\textbf{0.998} \\
\bottomrule
\end{tabular}}
\end{table*}

\subsection{Main Results}
\label{sec:main_results}
Table~\ref{tab:main_grid} reports the full grid: 17 benchmarks under all seven models.
On the three OpenAI models PairCoder improves 44 of the 48 cells; the four exceptions are three
exact ties at saturated oracles (ChartMimic execution at \texttt{gpt-5.4-mini}, BigCodeBench at
\texttt{gpt-5.5}, and StarVector rendering at \texttt{gpt-5.5}) and one small regression
(Plot2Code execution at \texttt{gpt-5.5}, 0.985 to 0.977, where the baseline already runs near
the ceiling and SSIM still improves, see Sec.~\ref{sec:multimodal}). Three patterns stand out. First, the largest gains concentrate
where the toolchain provides the richest signal: 3DCodeBench executability rises by up to 58
points, DaTikZ compile rate by 10 to 30 points, and WebApp1K and RTLLM each by up to 22 points. Second, gains often grow with model capability rather than shrink: HumanEval-X
C++ improves by 8.0, 11.0, and 31.1 points on \texttt{gpt-5.4-mini}, \texttt{gpt-5.4}, and
\texttt{gpt-5.5} respectively, because a stronger Navigator reviews more accurately and a
stronger Driver exploits the feedback better. Third, on benchmarks whose baseline is already
near the ceiling the loop terminates immediately via \texttt{[NOERROR]} and safely ties
instead of regressing, which is the designed behavior of the conservative review protocol.

\subsection{Cross-Vendor Generality}
\label{sec:cross_vendor}
The four right columns of Table~\ref{tab:main_grid} repeat the study on models from two
further vendors. The headline largely transfers: artifact domains with a strong toolchain signal improve on
nearly every applicable model. DaTikZ gains 10 to 30 compile points on all four models, 3DCodeBench gains
22 to 33 executability points on three of four, WebApp1K gains 2.5 to 22.5 points wherever
the model can write React at all, and the vision capable \texttt{doubao-seed-2.0-mini}
improves every chart, SVG, and CAD metric. The same columns also delimit the method honestly:
on benchmarks where the verification signal is weak (DS-1000 signature execution without
functional tests, BigCodeBench library calls, RTLLM compile only evidence), the weaker or non
GPT navigators occasionally churn working code and small regressions appear (for example
DS-1000 on the DeepSeek models). This is exactly the boundary in the qualitative pattern
above: where the toolchain cannot adjudicate, review quality becomes model bound; where it
can, gains appear across all vendors; Appendix~\ref{app:crossmodel} (Figures~\ref{fig:xm_chart}--\ref{fig:xm_tdcb}) shows the same tasks rendered across these models side by side. As a negative control we also ran GeoCodeBench~\cite{li2026benchmarking}, a PhD level
3D vision benchmark on which both arms score at the floor (about 0.02), and PairCoder ties
rather than regresses.

\subsection{Generalization to Parametric 3D Code (P3D-Bench)}
\label{sec:p3dbench}
To test whether the verification-grounded loop transfers to a target where the
artifact is a \emph{3D solid} rather than a program output, we evaluate on
P3D-Bench~\cite{p3dbench2026}, a benchmark in which the model writes a parametric
construction program that is compiled to a mesh and scored against ground-truth geometry.
P3D-Bench spans three tracks (text-to-3D, image-to-3D, and assembly-3D) across several
program formats (minimal-JSON, OpenSCAD, CadQuery, Three.js). We evaluate the \emph{text-to-3D}
track primarily in the minimal-JSON format, the one configuration whose ground truth is
materializable from the openly licensed Text2CAD~v1.1 corpus, and additionally report executable
validity for the OpenSCAD format. For the image-to-3D and assembly-3D tracks, whose full
evaluation sets build on the licensed Fusion~360 Gallery, we report a three-case in-repo
demonstration (Appendix~\ref{app:p3d}, Table~\ref{tab:p3d_demo}), on which PairCoder improves
validity, geometry, and topology in the CadQuery format of both tracks and ties on Three.js,
where the baseline already compiles all three cases.\footnote{We report P3D-Bench's official metrics:
executable validity, geometry (Chamfer, F-score, normal consistency, IoU), and topology.}
The Navigator's verification predicate $\psi$ is simply the benchmark's own compiler (``does
the program build a valid solid?''), and we report the official metrics: executable validity, a
geometry score (Chamfer distance, F-score, normal consistency and volumetric IoU, combined as the
worst-filled composite that zeroes invalid programs), and a topology score. We run the full
400-case text-to-3D split with thinking disabled.

The P3D-Bench rows of Table~\ref{tab:main_grid} show the same headroom-tracking
pattern as the rest of the grid: PairCoder improves \emph{every} metric on
\emph{every} model with no regressions. The weakest baseline
(\texttt{doubao-1.5-lite}, $14.0\%$ valid) gains $+28.7$ points of validity,
$+0.145$ geometry and $+0.280$ topology, and \texttt{gpt-5.4} ($71.5\%$) gains
$+27.5/+0.136/+0.275$, while near-saturated models such as \texttt{gpt-5.4-mini}
($97.3\%$) and \texttt{deepseek-v4-flash} ($94.5\%$) still improve rather than
regress. Crucially, on programs the baseline already compiles the geometry is
unchanged: PairCoder intervenes only when the artifact fails to build, so coverage
rises without disturbing correct solids, the conservative-review behavior of
Sec.~\ref{sec:main_results}, now measured on 3D geometry and topology.
Appendix~\ref{app:p3d} breaks the geometry and topology columns into their official sub-metrics
(capped Chamfer, F-score@0.05/0.01, normal consistency, and volumetric IoU for geometry,
Table~\ref{tab:p3d_detail}; no-open-edge, inverted-normal, and non-manifold ratios for topology,
Table~\ref{tab:p3d_topo}): PairCoder
improves every sub-metric on every model, while on the programs the single model already compiles
PairCoder keeps that program unchanged, so its geometry is \emph{identical} there and the
Chamfer distance on that common set matches the baseline exactly. This makes the source of the
gain unambiguous: it is coverage of non-executable programs, not re-optimization of already-valid
geometry: PairCoder repairs programs the single model fails to build at all into valid solids.
Appendix~\ref{app:p3d} additionally illustrates a vision-grounded geometry-refinement variant that
improves already-valid but inaccurate solids (Fig.~\ref{fig:p3d_improved}), for example raising
F-score@0.05 from $0.23$ to $0.95$. A second program format tells the same story: with OpenSCAD
in place of minimal-JSON, PairCoder drives validity to a perfect $1.000$ on all three models
tested, though the gain is smaller because the baseline is already more fluent in OpenSCAD
(Appendix~\ref{app:p3d}, Table~\ref{tab:p3d_openscad}).

\subsection{Token Economics}
\label{sec:tokens}
\begin{figure}[htbp]
    \centering
    \includegraphics[width=0.94\linewidth]{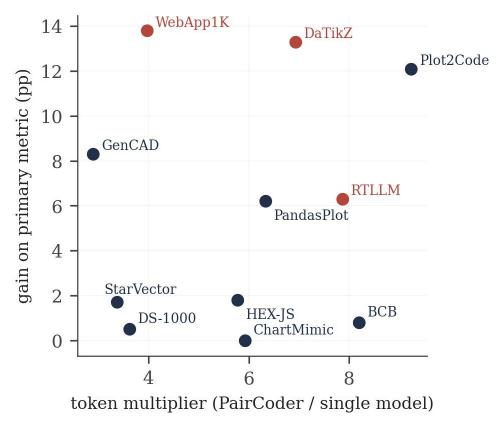}
    \caption{Cost versus benefit at \texttt{gpt-5.4-mini}: per-benchmark token multiplier against
    primary-metric gain; colours follow the families of Fig.~1.}
    \label{fig:cost_gain}
\end{figure}

We log every token in both arms. Across the full grid PairCoder consumes 2.9$\times$ to
9.2$\times$ the tokens of the single model baseline depending on the benchmark (for example
2.9$\times$ on GenCAD-Code, 3.4$\times$ on StarVector, 4.0$\times$ on WebApp1K, 5.9$\times$ on
ChartMimic, 6.9$\times$ on DaTikZ, 7.9$\times$ on RTLLM where simulation failures trigger the
most revision rounds, and 9.2$\times$ on Plot2Code),
with an overall ratio of 7.4$\times$ over 138M logged tokens (the per benchmark breakdown is in Table~\ref{tab:quant_tokens}). This is the cost of the review
rounds themselves: tasks accepted in the first round cost roughly 2$\times$ a single pass
(one generation plus one review), and the multiplier grows only when the Navigator actually
finds errors, that is, exactly when the additional spend converts into accuracy
(Fig.~\ref{fig:cost_gain}). Team style multi agent frameworks reported in prior work consume an order of magnitude more
than a single call; while we do not re-run those systems on our benchmarks, PairCoder's
measured 3 to 9 times envelope is well below that order of magnitude overhead.

\subsection{Role Switching Analysis}
\label{sec:roleswitch}
\begin{figure}[htbp]
\centering
\includegraphics[width=\columnwidth]{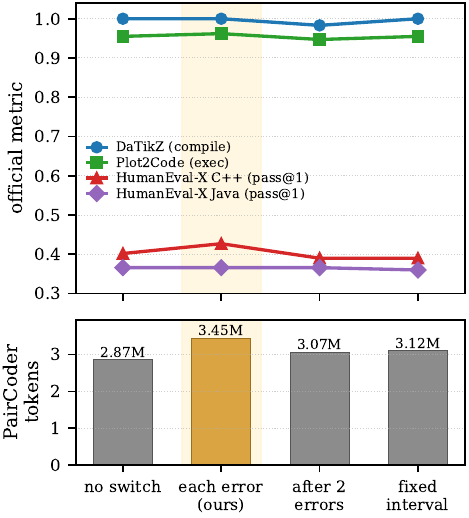}
\caption{Role-switching policy at \texttt{gpt-5.4-mini}: official metric (top) and PairCoder
token cost (bottom) for four policies; see Sec.~\ref{sec:roleswitch}.}
\label{fig:roleswitch}
\end{figure}

The error triggered role switch is a design choice; here we test it directly at
\texttt{gpt-5.4-mini} on four benchmarks, against three alternatives at matched budget:
\emph{no switch} (the Driver keeps the keyboard, the Navigator always reviews),
\emph{switch on each error} (the Navigator that just flagged an error takes over immediately),
\emph{switch after two consecutive errors}, and a \emph{fixed interval} switch every two
rounds regardless of the signal. Figure~\ref{fig:roleswitch} reports the official metric and the token cost for
each policy.

Switching on each error is the best or tied-best policy on all four benchmarks, and never
the worst. The effect is small where the loop is short (DaTikZ saturates near a perfect
compile rate under every policy) but clear where iteration matters: on HumanEval-X C++,
switching immediately reaches 0.427 while delaying the switch to two consecutive errors drops
to 0.390, the same as never switching, because the agent that diagnosed the fault is the one
best positioned to fix it, and keeping it in the reviewer seat wastes that diagnosis.
No switch and the fixed interval are competitive but consistently a step behind. We therefore
adopt switch on each error as the default in all other experiments. This updates the earlier
preference for a two error threshold, which we had observed on Python only; on the broader
benchmark set immediate switching is the more reliable choice.

\subsection{Effect of Reasoning Effort}
\label{sec:think}
The main results disable model thinking. We additionally test whether enabling reasoning
changes the picture, on the three Python benchmarks where the no-thinking gains are smallest,
at \texttt{gpt-5.4-mini}. Enabling high reasoning effort raises both arms' baselines, as
expected, and PairCoder still helps where the verification signal is informative: on
BigCodeBench the gain grows from $+0.8$ points (no thinking, 0.450 to 0.458) to $+5.0$ points
(high, 0.425 to 0.475), because a stronger Driver produces candidates whose remaining errors
are more often caught by the Navigator's doctest evidence. On DS-1000, whose only signal is
that a function executes on its signature (no functional tests), thinking lifts the baseline
from 0.250 to 0.310 but PairCoder neither helps nor hurts beyond noise (0.310 to 0.300),
consistent with the qualitative pattern above: where the toolchain cannot adjudicate correctness,
extra reasoning improves the Driver but gives the Navigator nothing new to verify. Reasoning
roughly triples per-call token cost, so we use it only for this analysis and report the main
grid with thinking disabled. (We omit hardware RTL from this comparison: at high effort the
small model spends most of its budget on reasoning and frequently truncates the actual Verilog
module, an artifact of the model rather than of the collaboration.)

\section{Code-Driven Multimodal Generation}
\label{sec:multimodal}

The central claim of this work is that pair programming turns code driven creation, not just
program synthesis, into a verified process. We test it on seven public benchmarks whose
outputs are executable artifacts in other modalities: scientific figures
(DaTikZ~\cite{belouadi2024automatikz}: caption to Ti\textit{k}Z/\LaTeX), charts
(Plot2Code~\cite{wu2024plot2code}, PandasPlotBench~\cite{galimzyanov2024drawing},
ChartMimic~\cite{yang2024chartmimic}: image or data to matplotlib), vector graphics
(StarVector~\cite{rodriguez2024starvector}: image to SVG), parametric CAD
(GenCAD-Code~\cite{alam2024gencad}: image to CadQuery), and 3D scenes
(3DCodeBench~\cite{gao20263dcodebench}: text to Blender script). In each domain the
Navigator's review is grounded in the same verification predicates $\psi_i$ as in
Sec.~\ref{RoleAssignment}: the candidate program is compiled, executed, or rendered, and the
result (including the rendered image for multimodal tasks) is supplied to the Navigator as
review evidence before it issues \texttt{[NOERROR]} or a concrete revision request. We report
each benchmark's official metric suite, namely execution or compile rate plus visual
similarity (SSIM~\cite{wang2004image}, CLIP~\cite{radford2021learning},
DINO~\cite{oquab2023dinov2}, SigLIP-2~\cite{tschannen2025siglip}) and geometric fidelity
(Chamfer distance), rather than execution success alone.

\begin{table}[htbp]
\caption{Multimodal code generation with \texttt{gpt-5.4-mini} (thinking off); each cell is
single model $\rightarrow$ \textbf{PairCoder} ($\downarrow$ better when lower).}
\label{tab:multimodal}
\centering
\scriptsize
\setlength{\tabcolsep}{3pt}
\resizebox{\columnwidth}{!}{%
\begin{tabular}{llc}
\toprule
\textbf{Benchmark} & \textbf{Metric} & \textbf{single $\rightarrow$ PairCoder} \\
\midrule
\multirow{2}{*}{DaTikZ (60)}
 & compile rate $\uparrow$ & 0.500 $\rightarrow$ \textbf{0.633} \\
 & SSIM / CLIP / DINO $\uparrow$ & .215/.327/.303 $\rightarrow$ \textbf{.256/.384/.338} \\
\midrule
\multirow{2}{*}{Plot2Code (132)}
 & execution rate $\uparrow$ & 0.841 $\rightarrow$ \textbf{0.962} \\
 & SSIM / CLIP $\uparrow$ & .412/.802 $\rightarrow$ \textbf{.474/.912} \\
\midrule
\multirow{2}{*}{PandasPlotBench (175)}
 & execution rate $\uparrow$ & 0.789 $\rightarrow$ \textbf{0.851} \\
 & SSIM / CLIP $\uparrow$ & .423/.728 $\rightarrow$ \textbf{.461/.794} \\
\midrule
\multirow{2}{*}{ChartMimic (60)}
 & execution rate $\uparrow$ & 0.967 $\rightarrow$ 0.967 \\
 & SSIM / CLIP $\uparrow$ & .578/.871 $\rightarrow$ .578/.861 \\
\midrule
\multirow{2}{*}{StarVector (60)}
 & render rate $\uparrow$ & 0.983 $\rightarrow$ \textbf{1.000} \\
 & SSIM / CLIP / DINO $\uparrow$ & .784/.929/.874 $\rightarrow$ \textbf{.801/.944/.885} \\
\midrule
\multirow{2}{*}{GenCAD-Code (60)}
 & execution rate $\uparrow$ & 0.900 $\rightarrow$ \textbf{0.983} \\
 & Chamfer (aggregate) $\downarrow$ & 0.233 $\rightarrow$ \textbf{0.166} \\
\midrule
\multirow{3}{*}{3DCodeBench (60)}
 & executability $\uparrow$ & 0.200 $\rightarrow$ \textbf{0.783} \\
 & SigLIP-2 / DINO (agg.) $\uparrow$ & .181/.101 $\rightarrow$ \textbf{.702/.376} \\
 & Chamfer (aggregate) $\downarrow$ & 4.85 $\rightarrow$ \textbf{2.62} \\
\bottomrule
\end{tabular}}
\end{table}

\begin{figure*}[t]
\centering
\begin{subfigure}{0.485\textwidth}\centering
  \includegraphics[width=\linewidth]{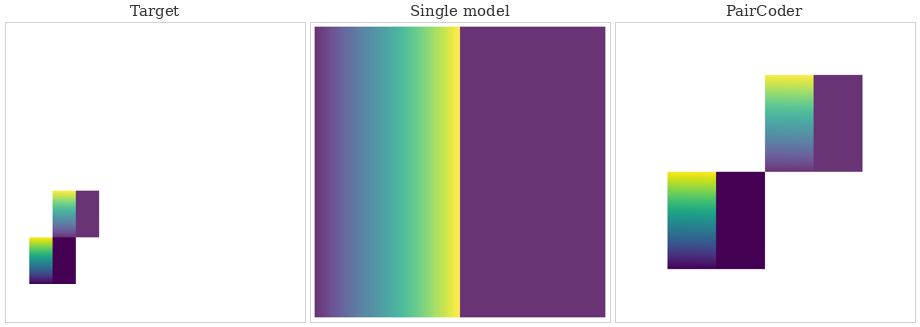}\caption{Plot2Code: image to matplotlib}\end{subfigure}\hfill
\begin{subfigure}{0.485\textwidth}\centering
  \includegraphics[width=\linewidth]{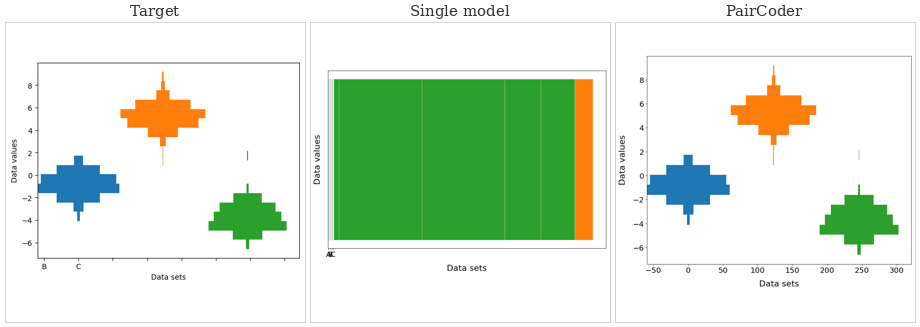}\caption{PandasPlotBench: data to matplotlib}\end{subfigure}\\[2pt]
\begin{subfigure}{0.485\textwidth}\centering
  \includegraphics[width=\linewidth]{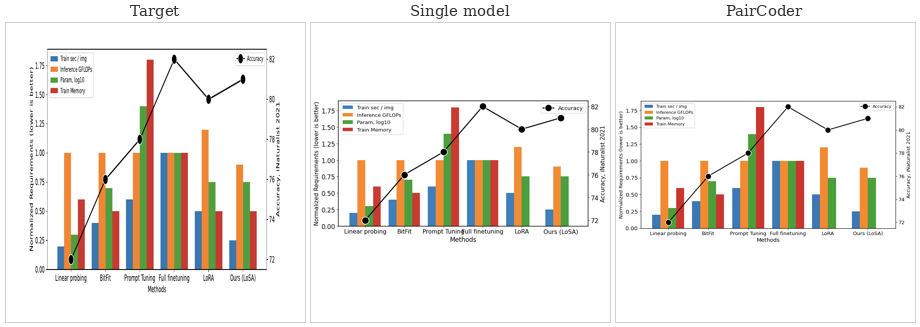}\caption{ChartMimic: chart to matplotlib}\end{subfigure}\hfill
\begin{subfigure}{0.485\textwidth}\centering
  \includegraphics[width=\linewidth]{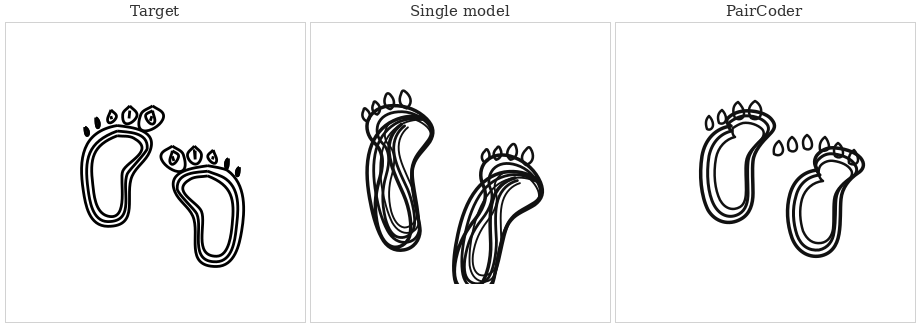}\caption{StarVector: image to SVG}\end{subfigure}\\[2pt]
\begin{subfigure}{0.485\textwidth}\centering
  \includegraphics[width=\linewidth]{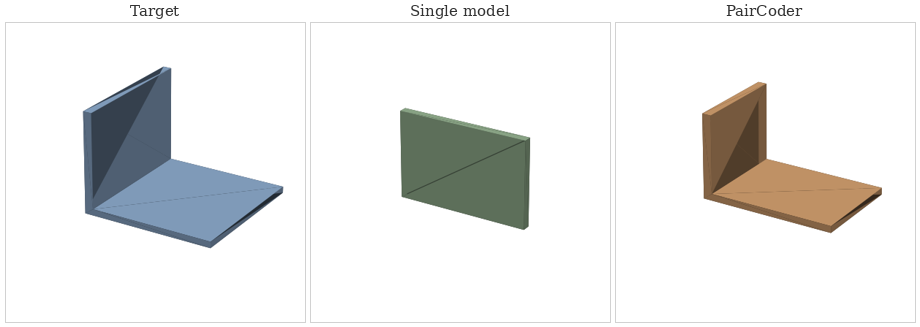}\caption{GenCAD-Code: image to CadQuery}\end{subfigure}\hfill
\begin{subfigure}{0.485\textwidth}\centering
  \includegraphics[width=\linewidth]{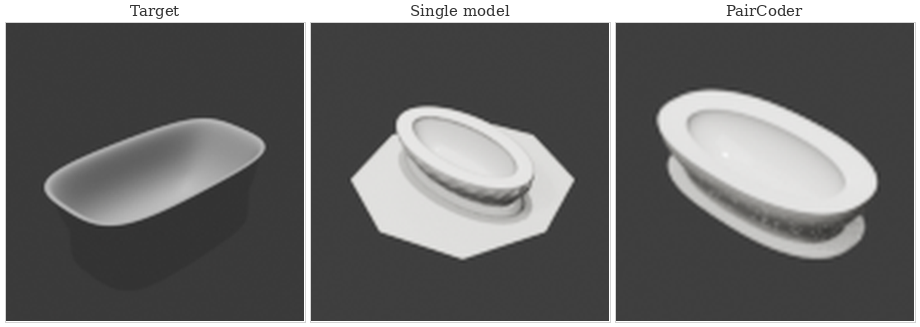}\caption{3DCodeBench: text to Blender}\end{subfigure}
\caption{Qualitative comparison across six artifact domains: target/reference (left), single
model (middle), PairCoder (right).}
\label{fig:multimodal_qual}
\end{figure*}

\paragraph{Quantitative results.}
Table~\ref{tab:multimodal} summarizes the \texttt{gpt-5.4-mini} results. PairCoder improves
every official metric on six of the seven benchmarks: not only are far more generated programs
executable, the artifacts they produce are also closer to the reference in pixel structure
(SSIM), semantics (CLIP, SigLIP-2, DINO), and geometry (Chamfer). The only exception is
ChartMimic, where the baseline already executes near the ceiling (about 0.97 to 1.0) and the
loop ties; notably, at \texttt{gpt-5.4} the same benchmark turns positive (execution 0.92 to
0.98, SSIM 0.53 to 0.58), confirming that the tie reflects a saturated oracle rather than a
limit of the collaboration. These gains transfer across vendors: with the volcano engine
models, DaTikZ improves on all four (doubao-1.5-lite $+30.0$, doubao-seed-2.0-mini $+23.3$,
deepseek-v3.2 $+23.4$, deepseek-v4-flash $+10.0$ compile points), 3DCodeBench gains
$+21.6$/$+33.3$/$+30.0$ executability points, and the multimodal doubao-seed-2.0-mini improves
nearly every chart, SVG, and CAD metric (for example, PandasPlotBench execution 0.91 to 0.99).

\paragraph{Qualitative comparison.}
Figure~\ref{fig:multimodal_qual} illustrates the dominant failure repair pattern behind these
numbers. Single model generations in these domains frequently die at the toolchain boundary:
\LaTeX{} that does not compile, matplotlib scripts that raise exceptions, Blender scripts that
produce no geometry. PairCoder converts a large fraction of these hard failures because the
Navigator receives the concrete compiler or runtime error (and, for image conditioned tasks, a
rendering of the current candidate next to the target) and returns an actionable critique;
after the error triggered role switch, the agent that diagnosed the fault repairs it directly.
When the baseline already renders, the same visual review instead nudges layout, proportions,
and styling toward the target, which is reflected in the consistent SSIM, CLIP, and DINO
gains. Appendix~\ref{app:gallery} expands this into one page per benchmark, with galleries
(Figures~\ref{fig:gal_plot2code1}--\ref{fig:gal_tdcb1}) and the full measured metric suites
(Tables~\ref{tab:quant_plot2code}--\ref{tab:quant_tdcb}); Appendix~\ref{example_paircoder}
walks through complete multi-round traces (Cases~\ref{app:case1}--\ref{app:case4}).

\paragraph{Cost.}
The collaboration costs about 3 to 9 times the tokens of single model inference in these
domains (for example, 2.9$\times$ on GenCAD-Code, 3.4$\times$ on StarVector, 5.9$\times$ on
ChartMimic, 6.9$\times$ on DaTikZ, and 9.2$\times$ on Plot2Code, measured per arm), with an
overall ratio of about 7$\times$ across the full grid; review rounds terminate early via
\texttt{[NOERROR]} on the majority of tasks that succeed immediately.

\section{Conclusion}
Code is increasingly how language models make things, from figures, charts, and vector
graphics to CAD parts, 3D scenes, and circuits, and in this regime the decisive failures occur
at the boundary between the program and the toolchain that renders it. We presented PairCoder,
a minimal two agent framework that closes this boundary: a Driver writes the program, a
Navigator reviews it with verdicts grounded in compiler, execution, and rendering evidence,
and error triggered role switching keeps the collaboration moving. Across 17 public
benchmarks, seven models, and three vendors, PairCoder improves essentially every verifiable
benchmark on its entire official metric suite, covering executability, visual similarity, and
geometric fidelity alike, while remaining within 2.9 to 9.2 times a single pass in cost (about 7 times overall), far below
team simulation systems; on classical synthesis it remains competitive with and often improves
strong single model inference (for example LiveCodeBench pass@1 from 0.94 to 0.99). The
broader lesson is a qualitative pattern for collaborative generation: gains concentrate
where the artifact is verifiable and the baseline leaves headroom, and fade where the oracle
is weak. Wherever a trustworthy toolchain can be made
to speak, and in code driven creation it almost always can, pair programming converts that
signal into reliably better artifacts, making it a practical recipe for trustworthy code
mediated generation of structured data and multimodal content wherever such an oracle exists.

\section*{Limitations}
While PairCoder improves code generation accuracy, it still requires substantially more tokens and wall clock time than single model inference, with about seven times the cost per task on average in our experiments (2.9 to 9.2 times depending on the benchmark). Although one PairCoder run can reach accuracy levels that otherwise require several retries, the added latency and token budget still constrain practical deployment. Reducing this cost while preserving accuracy remains an important direction for future work.

\bibliography{custom}

\newpage

\clearpage

\appendix

\section{Multi-Round PairCoder Traces on Code-Driven Multimodal Benchmarks}\label{example_paircoder}

This appendix documents complete PairCoder sessions on the multimodal benchmarks of
Sec.~\ref{sec:multimodal}. Each trace shows the Driver's program, the verification evidence
$\psi$ attached to the Navigator's review (execution status and, for image conditioned tasks,
the structural similarity of the current rendering to the target), the Navigator's verbatim
verdict, and the error triggered role switches. Renders of the rounds accompany each trace so
the effect of each review can be inspected visually. Traces use \texttt{gpt-5.4-mini} unless
stated otherwise; prompts and the review protocol are identical to the main experiments.

\subsection{Case 1 (Plot2Code): review, role switch, and cross acceptance}
\label{app:case1}
\begin{figure}[htbp]
    \centering
    \includegraphics[width=\linewidth]{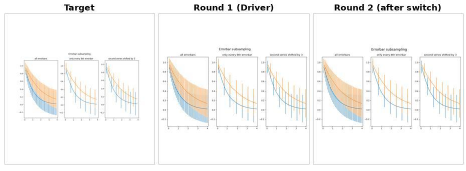}
    \caption{Case 1 rendering progression: target (left), the Driver's first program
    (middle), the program after the role switch (right).}
    \label{fig:trace_p2c33}
\end{figure}

The Driver reproduces a three panel errorbar chart from the target image
(Fig.~\ref{fig:trace_p2c33}). Its first program
already runs and is close to the target, but the Navigator, looking at the current rendering
next to the target, finds a concrete visual deviation and rejects the round. After the role
switch the former Navigator takes the keyboard and applies its own fix, and the former
Driver, now reviewing, accepts with \texttt{[NOERROR]}. The session costs two generations and
two reviews.

\begin{messagebox}{driverColor}{Round 1, Driver (initial program, excerpt)}
\begin{lstlisting}[language=Python]
fig, axs = plt.subplots(1, 3,
    figsize=(12, 6), sharex=True)
fig.suptitle("Errorbar subsampling",
    fontsize=14)
axs[0].errorbar(x, y1, yerr=yerr, ...)
axs[1].errorbar(x, y1, yerr=yerr,
    errorevery=6)
axs[2].errorbar(x, y1, yerr=yerr,
    errorevery=(6, 3))
\end{lstlisting}
\end{messagebox}

\begin{messagebox}{systemColor}{Verification evidence $\psi$ shown to the Navigator}
\begin{lstlisting}[language=Python]
runs OK; structural similarity to
the target chart = 0.876
\end{lstlisting}
\end{messagebox}

\begin{messagebox}{navigatorColor}{Round 1, Navigator review (verbatim)}
\begin{lstlisting}[language=Python]
The code has a visual mismatch with
the target figure:
- Line causing the mismatch:
  fig.suptitle("Errorbar subsampling",
               fontsize=14)
- Why: the target image uses a larger,
  more prominent figure title and
  slightly different top spacing. With
  fontsize=14 the title is too small
  compared to the reference.
- Fix: increase the suptitle size and
  adjust its vertical placement.
\end{lstlisting}
\end{messagebox}

\begin{messagebox}{navigatorColor}{Round 2, after role switch: former Navigator writes the fix (diff)}
\begin{lstlisting}[language=Python]
- fig.suptitle("Errorbar subsampling",
-     fontsize=14)
+ fig.suptitle("Errorbar subsampling",
+     fontsize=18, y=0.98)
\end{lstlisting}
\end{messagebox}

\begin{messagebox}{driverColor}{Round 2, former Driver now reviews}
\begin{lstlisting}[language=Python]
[NOERROR]
\end{lstlisting}
\end{messagebox}

\subsection{Case 2 (Plot2Code): review oscillation and the quality fallback}
\label{app:case2}
\begin{figure}[htbp]
    \centering
    \includegraphics[width=\linewidth]{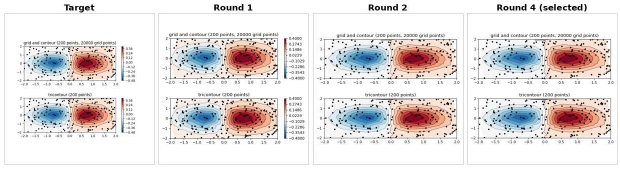}
    \caption{Case 2 rendering progression: target, rounds 1 and 2, and the round selected by
    the quality fallback after budget exhaustion.}
    \label{fig:trace_p2c24}
\end{figure}

A harder session on a two panel contour task. The pair improves the program over three
revision rounds (contour levels, then colorbars), but the two reviewers disagree about
whether the target contains colorbars, so no round is accepted within the budget. PairCoder
then falls back to the quality criterion of Algorithm~\ref{alg:paircoder} and returns the
round with the highest verified similarity score, which is also the visually closest
rendering (Fig.~\ref{fig:trace_p2c24}). The fallback turns an unresolved review debate into a
safe outcome instead of a failure.

\begin{messagebox}{navigatorColor}{Round 1, Navigator review (verbatim, trimmed)}
\begin{lstlisting}[language=Python]
The code has a definite mismatch:
- Problem line:
  levels = np.linspace(-0.48, 0.40, 15)
- Why: the prompt says the plots should
  use 14 levels, and the target image's
  contour/colorbar scaling does not
  match this fixed custom range.
- Concrete fix: use 14 contour levels
  directly in both plots.
\end{lstlisting}
\end{messagebox}

\begin{messagebox}{driverColor}{Round 2, former Driver now reviews (verbatim, trimmed)}
\begin{lstlisting}[language=Python]
Missing colorbars.
- The target figure shows a colorbar
  beside each subplot, but the code
  never adds one. The exact missing
  lines are:
  fig.colorbar(cf, ax=ax)
  fig.colorbar(cf2, ax=ax)
\end{lstlisting}
\end{messagebox}

\begin{messagebox}{navigatorColor}{Round 3, the other agent disagrees (verbatim, trimmed)}
\begin{lstlisting}[language=Python]
The code still has a concrete mismatch:
- Problem lines: fig.colorbar(cf, ax=ax)
  and fig.colorbar(cf2, ax=ax)
- Why: the target image shows NO
  colorbars. Adding them changes the
  layout and shrinks the axes.
\end{lstlisting}
\end{messagebox}

\subsection{Case 3 (DaTikZ, \texttt{doubao-1.5-lite}): compile error repair on a weak model}
\label{app:case3}
\begin{figure}[htbp]
    \centering
    \includegraphics[width=\linewidth]{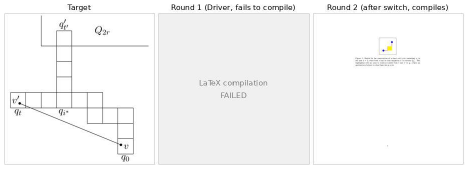}
    \caption{Case 3 rendering progression on \texttt{doubao-1.5-lite}: target (left), the
    Driver's first (non-compiling) program (middle), the repaired program after the role switch (right).}
    \label{fig:trace_datikz17}
\end{figure}

The same protocol on the weakest model in our pool, where PairCoder lifts the DaTikZ compile
rate from 0.550 to 0.850. The Driver's first Ti\textit{k}Z program fails to compile because
its caption references an undefined cross reference macro. The Navigator receives the
compiler verdict as evidence, identifies the exact offending construct, and proposes the
minimal removal; after the role switch the repaired program compiles and the former Driver
accepts (Fig.~\ref{fig:trace_datikz17}).

\begin{messagebox}{systemColor}{Verification evidence $\psi$ shown to the Navigator}
\begin{lstlisting}[language=Python]
LaTeX compilation FAILED
\end{lstlisting}
\end{messagebox}

\begin{messagebox}{navigatorColor}{Round 1, Navigator review (verbatim, trimmed)}
\begin{lstlisting}[language=Python]
The error is that the code uses
\cref{claim:paths} which is not defined
anywhere in the provided code. To fix
this, either define the claim:paths
macro appropriately in the preamble or
remove the reference from the caption.
Specific fix: remove the
\cref{claim:paths} part from the
caption.
\end{lstlisting}
\end{messagebox}

\begin{messagebox}{driverColor}{Round 2, former Driver now reviews}
\begin{lstlisting}[language=Python]
[NOERROR]
\end{lstlisting}
\end{messagebox}

\subsection{Case 4 (DaTikZ): first round acceptance, the cheap path}
\label{app:case4}
For completeness, the most common trace shape: the Driver's first program compiles, the
Navigator receives the passing evidence, finds no concrete error, and accepts immediately
with \texttt{[NOERROR]}. The session costs one generation plus one review, roughly twice a
single pass, which is why the average token multiplier in Sec.~\ref{sec:tokens} stays in the
3 to 9 range even though failing tasks iterate several times.

\begin{messagebox}{systemColor}{Verification evidence $\psi$}
\begin{lstlisting}[language=Python]
compiles and renders OK
\end{lstlisting}
\end{messagebox}

\begin{messagebox}{navigatorColor}{Round 1, Navigator}
\begin{lstlisting}[language=Python]
[NOERROR]
\end{lstlisting}
\end{messagebox}

\section{Per-Benchmark Qualitative and Quantitative Results}
\label{app:gallery}
This appendix devotes one full page to each artifact benchmark: a gallery of qualitative
comparisons on top and that benchmark's complete measured metric suite below. In every gallery
triplet the target or ground truth rendering is on the left, the single model baseline's
artifact is in the middle, and PairCoder's artifact is on the right (copper border), with the
task id and input quoted beneath; each example is a distinct task. For the chart, plot, and
vector benchmarks every example has both arms rendering with PairCoder quantitatively closer to
the target, so the comparison isolates artifact quality. For DaTikZ, GenCAD-Code, and
3DCodeBench, where PairCoder's measured advantage is concentrated in executability, the
galleries lead with fidelity gains and then show executability repairs, in which the baseline
produces no artifact (gray panel) and PairCoder renders successfully. Each table reports every
metric we measured for that benchmark, as ``single model $\rightarrow$ PairCoder,'' with green
cells marking improvements and aggregate scores counting non rendering generations as zero.
\label{app:quant}

\begin{figure*}[p]
    \centering
    \includegraphics[width=\textwidth]{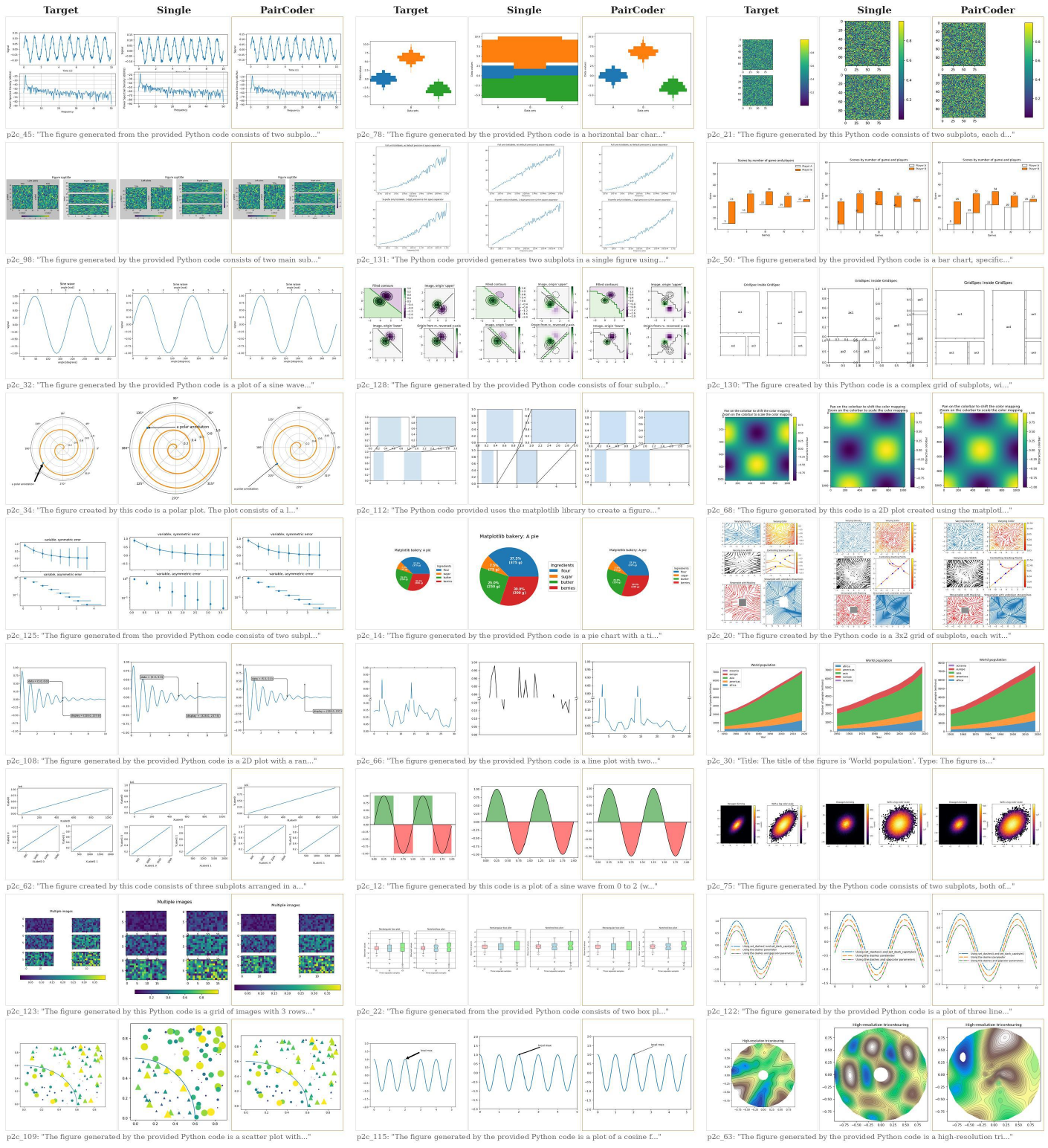}
    \captionof{figure}{Plot2Code qualitative gallery (27 distinct examples); both arms render and PairCoder is quantitatively closer to the target.}
    \label{fig:gal_plot2code1}

    \vspace{0.5em}
    \centering\scriptsize\setlength{\tabcolsep}{4pt}\renewcommand{\arraystretch}{1.05}
    \begin{tabular}{lccc}
\toprule
\textbf{Model} & exec $\uparrow$ & SSIM $\uparrow$ & CLIP $\uparrow$ \\
\midrule
gpt-5.4-mini & \cellcolor{improvebg}0.841 $\rightarrow$ \textbf{0.962} & \cellcolor{improvebg}0.412 $\rightarrow$ \textbf{0.474} & \cellcolor{improvebg}0.802 $\rightarrow$ \textbf{0.912} \\
gpt-5.4 & \cellcolor{improvebg}0.962 $\rightarrow$ \textbf{0.977} & \cellcolor{improvebg}0.476 $\rightarrow$ \textbf{0.495} & \cellcolor{improvebg}0.916 $\rightarrow$ \textbf{0.920} \\
gpt-5.5 & 0.985 $\rightarrow$ 0.977 & \cellcolor{improvebg}0.487 $\rightarrow$ \textbf{0.496} & 0.943 $\rightarrow$ 0.931 \\
doubao-seed-2.0-mini & \cellcolor{improvebg}0.909 $\rightarrow$ \textbf{0.917} & \cellcolor{improvebg}0.410 $\rightarrow$ \textbf{0.443} & 0.835 $\rightarrow$ 0.823 \\
\bottomrule
\end{tabular}
    \captionof{table}{Plot2Code: full measured metric suite for every applicable model (single model $\rightarrow$ \textbf{PairCoder}).}
    \label{tab:quant_plot2code}
\end{figure*}

\begin{figure*}[p]
    \centering
    \includegraphics[width=\textwidth]{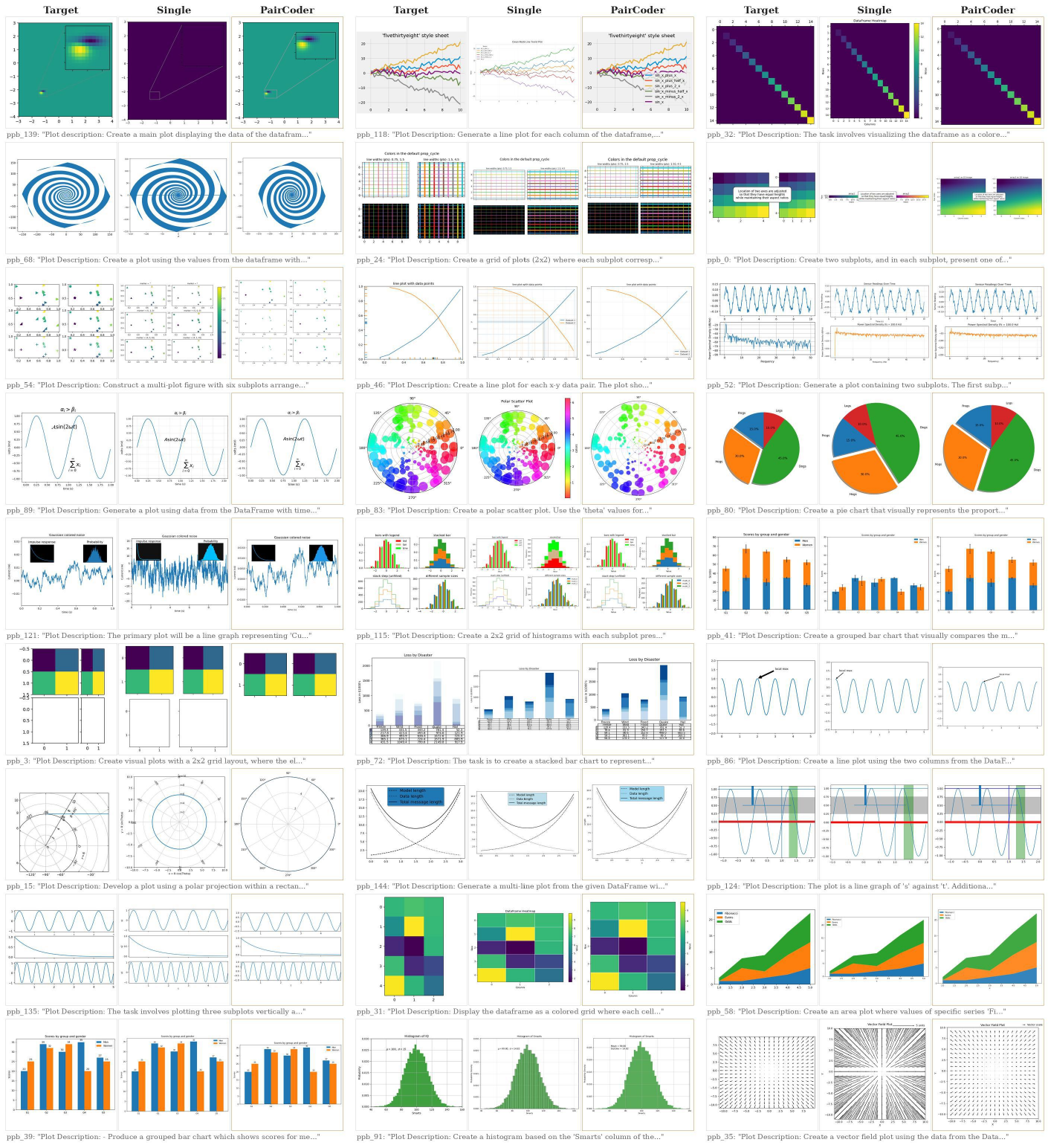}
    \captionof{figure}{PandasPlotBench qualitative gallery (27 distinct examples); both arms render and PairCoder is quantitatively closer to the target.}
    \label{fig:gal_ppb1}

    \vspace{0.5em}
    \centering\scriptsize\setlength{\tabcolsep}{4pt}\renewcommand{\arraystretch}{1.05}
    \begin{tabular}{lccc}
\toprule
\textbf{Model} & exec $\uparrow$ & SSIM $\uparrow$ & CLIP $\uparrow$ \\
\midrule
gpt-5.4-mini & \cellcolor{improvebg}0.789 $\rightarrow$ \textbf{0.851} & \cellcolor{improvebg}0.423 $\rightarrow$ \textbf{0.461} & \cellcolor{improvebg}0.728 $\rightarrow$ \textbf{0.794} \\
gpt-5.4 & \cellcolor{improvebg}0.966 $\rightarrow$ \textbf{0.994} & \cellcolor{improvebg}0.567 $\rightarrow$ \textbf{0.610} & \cellcolor{improvebg}0.909 $\rightarrow$ \textbf{0.944} \\
gpt-5.5 & \cellcolor{improvebg}0.960 $\rightarrow$ \textbf{0.977} & \cellcolor{improvebg}0.597 $\rightarrow$ \textbf{0.644} & \cellcolor{improvebg}0.907 $\rightarrow$ \textbf{0.942} \\
doubao-seed-2.0-mini & \cellcolor{improvebg}0.909 $\rightarrow$ \textbf{0.989} & \cellcolor{improvebg}0.491 $\rightarrow$ \textbf{0.534} & \cellcolor{improvebg}0.846 $\rightarrow$ \textbf{0.922} \\
\bottomrule
\end{tabular}
    \captionof{table}{PandasPlotBench: full measured metric suite for every applicable model (single model $\rightarrow$ \textbf{PairCoder}).}
    \label{tab:quant_ppb}
\end{figure*}

\begin{figure*}[p]
    \centering
    \includegraphics[width=\textwidth]{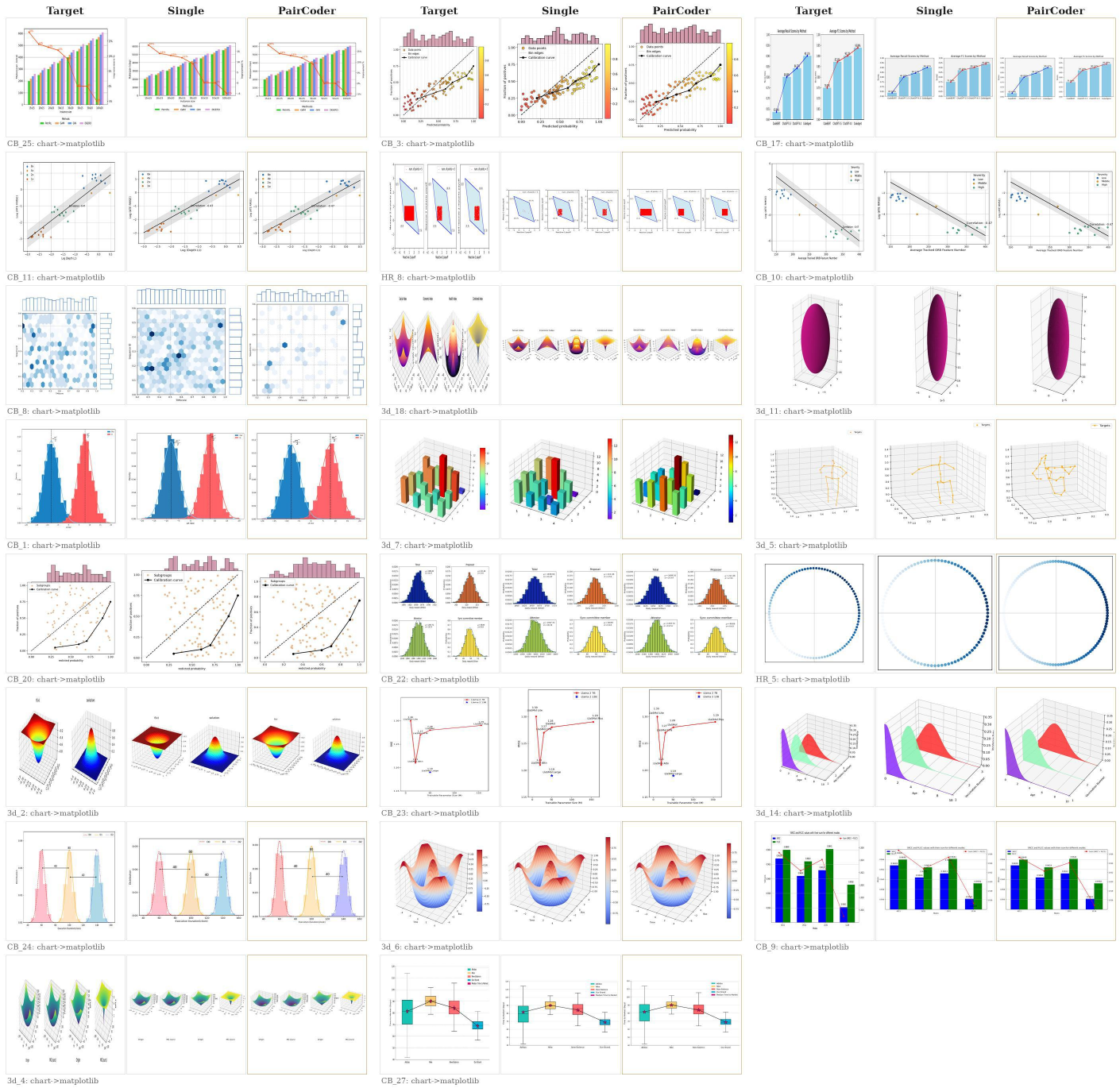}
    \captionof{figure}{ChartMimic qualitative gallery (23 distinct examples); both arms render and PairCoder is quantitatively closer to the target.}
    \label{fig:gal_chartmimic1}

    \vspace{0.5em}
    \centering\scriptsize\setlength{\tabcolsep}{4pt}\renewcommand{\arraystretch}{1.05}
    \begin{tabular}{lccc}
\toprule
\textbf{Model} & exec $\uparrow$ & SSIM $\uparrow$ & CLIP $\uparrow$ \\
\midrule
gpt-5.4-mini & 0.967 $\rightarrow$ 0.967 & \cellcolor{improvebg}0.578 $\rightarrow$ \textbf{0.578} & 0.871 $\rightarrow$ 0.861 \\
gpt-5.4 & \cellcolor{improvebg}0.917 $\rightarrow$ \textbf{0.983} & \cellcolor{improvebg}0.533 $\rightarrow$ \textbf{0.582} & \cellcolor{improvebg}0.820 $\rightarrow$ \textbf{0.848} \\
gpt-5.5 & \cellcolor{improvebg}0.983 $\rightarrow$ \textbf{1.000} & \cellcolor{improvebg}0.593 $\rightarrow$ \textbf{0.613} & \cellcolor{improvebg}0.874 $\rightarrow$ \textbf{0.887} \\
doubao-seed-2.0-mini & \cellcolor{improvebg}0.900 $\rightarrow$ \textbf{0.950} & \cellcolor{improvebg}0.443 $\rightarrow$ \textbf{0.476} & \cellcolor{improvebg}0.770 $\rightarrow$ \textbf{0.808} \\
\bottomrule
\end{tabular}
    \captionof{table}{ChartMimic: full measured metric suite for every applicable model (single model $\rightarrow$ \textbf{PairCoder}).}
    \label{tab:quant_chartmimic}
\end{figure*}

\begin{figure*}[p]
    \centering
    \includegraphics[width=\textwidth]{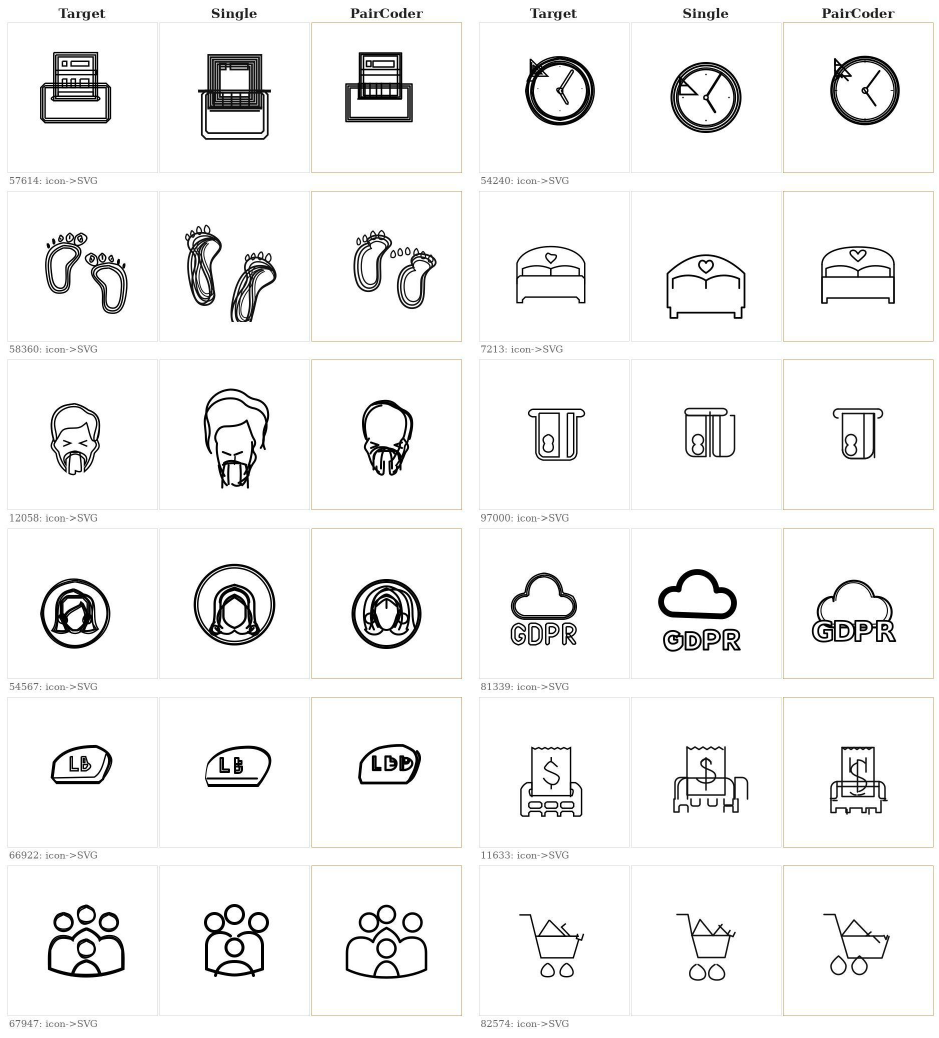}
    \captionof{figure}{StarVector qualitative gallery (12 distinct examples); both arms render and PairCoder is quantitatively closer to the target.}
    \label{fig:gal_starvector1}

    \vspace{0.5em}
    \centering\scriptsize\setlength{\tabcolsep}{4pt}\renewcommand{\arraystretch}{1.05}
    \begin{tabular}{lcccc}
\toprule
\textbf{Model} & render rate $\uparrow$ & SSIM $\uparrow$ & CLIP $\uparrow$ & DINO $\uparrow$ \\
\midrule
gpt-5.4-mini & \cellcolor{improvebg}0.983 $\rightarrow$ \textbf{1.000} & \cellcolor{improvebg}0.784 $\rightarrow$ \textbf{0.801} & \cellcolor{improvebg}0.929 $\rightarrow$ \textbf{0.944} & \cellcolor{improvebg}0.874 $\rightarrow$ \textbf{0.885} \\
gpt-5.4 & \cellcolor{improvebg}0.900 $\rightarrow$ \textbf{1.000} & \cellcolor{improvebg}0.776 $\rightarrow$ \textbf{0.873} & \cellcolor{improvebg}0.859 $\rightarrow$ \textbf{0.954} & \cellcolor{improvebg}0.820 $\rightarrow$ \textbf{0.906} \\
gpt-5.5 & 1.000 $\rightarrow$ 1.000 & \cellcolor{improvebg}0.777 $\rightarrow$ \textbf{0.787} & 0.961 $\rightarrow$ 0.960 & 0.938 $\rightarrow$ 0.934 \\
doubao-seed-2.0-mini & \cellcolor{improvebg}0.983 $\rightarrow$ \textbf{1.000} & \cellcolor{improvebg}0.758 $\rightarrow$ \textbf{0.770} & \cellcolor{improvebg}0.883 $\rightarrow$ \textbf{0.915} & \cellcolor{improvebg}0.793 $\rightarrow$ \textbf{0.819} \\
\bottomrule
\end{tabular}
    \captionof{table}{StarVector: full measured metric suite for every applicable model (single model $\rightarrow$ \textbf{PairCoder}).}
    \label{tab:quant_svg}
\end{figure*}

\begin{figure*}[p]
    \centering
    \includegraphics[width=\textwidth]{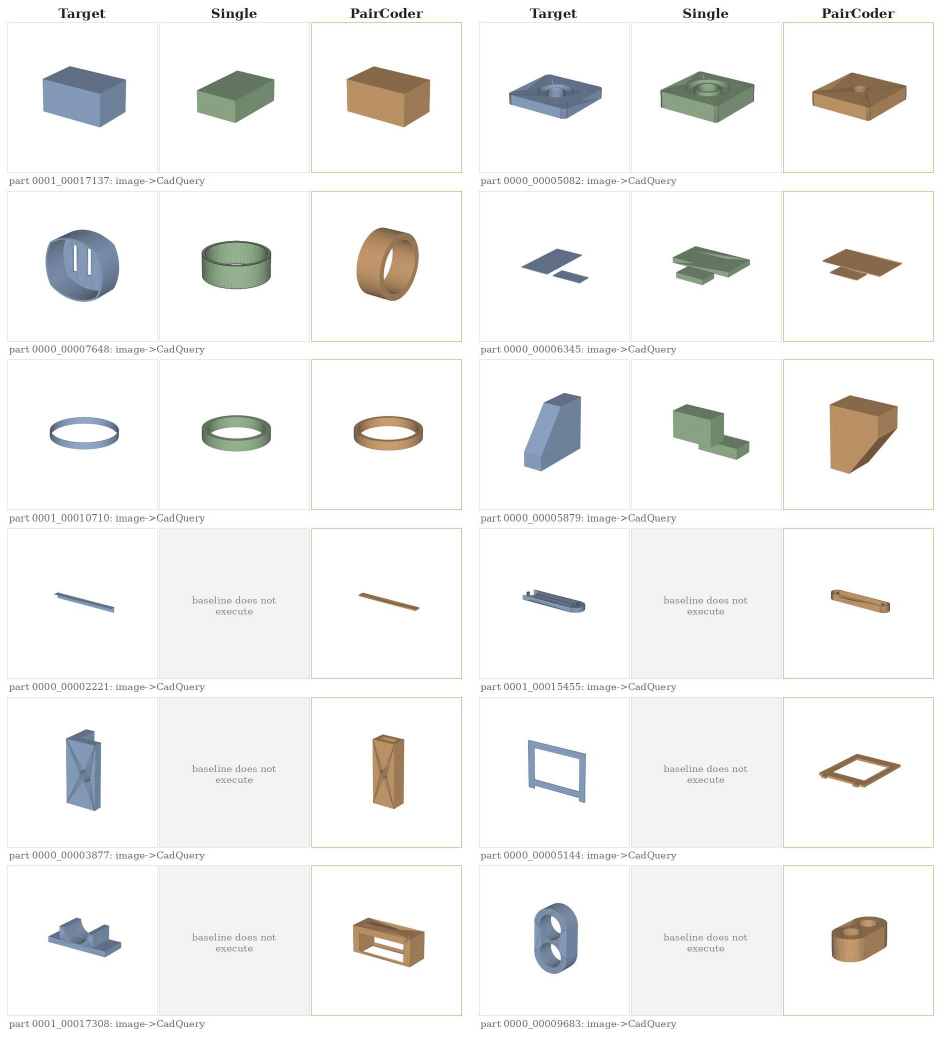}
    \captionof{figure}{GenCAD-Code qualitative gallery (12 distinct examples) demonstrating PairCoder's advantage; gray panels mark baselines that produce no artifact.}
    \label{fig:gal_gencad1}

    \vspace{0.5em}
    \centering\scriptsize\setlength{\tabcolsep}{4pt}\renewcommand{\arraystretch}{1.05}
    \begin{tabular}{lccc}
\toprule
\textbf{Model} & exec $\uparrow$ & Chamfer (cond.) $\downarrow$ & Chamfer (agg.) $\downarrow$ \\
\midrule
gpt-5.4-mini & \cellcolor{improvebg}0.900 $\rightarrow$ \textbf{0.983} & 0.148 $\rightarrow$ 0.152 & \cellcolor{improvebg}0.233 $\rightarrow$ \textbf{0.166} \\
gpt-5.4 & \cellcolor{improvebg}0.867 $\rightarrow$ \textbf{0.983} & \cellcolor{improvebg}0.145 $\rightarrow$ \textbf{0.140} & \cellcolor{improvebg}0.259 $\rightarrow$ \textbf{0.155} \\
doubao-seed-2.0-mini & \cellcolor{improvebg}0.833 $\rightarrow$ \textbf{0.967} & 0.169 $\rightarrow$ 0.185 & \cellcolor{improvebg}0.308 $\rightarrow$ \textbf{0.212} \\
gpt-5.5 & \cellcolor{improvebg}0.917 $\rightarrow$ \textbf{1.000} & \cellcolor{improvebg}0.143 $\rightarrow$ \textbf{0.126} & \cellcolor{improvebg}0.215 $\rightarrow$ \textbf{0.126} \\
\bottomrule
\end{tabular}
    \captionof{table}{GenCAD-Code: full measured metric suite for every applicable model (single model $\rightarrow$ \textbf{PairCoder}).}
    \label{tab:quant_gencad}
\end{figure*}

\begin{figure*}[p]
    \centering
    \includegraphics[width=\textwidth]{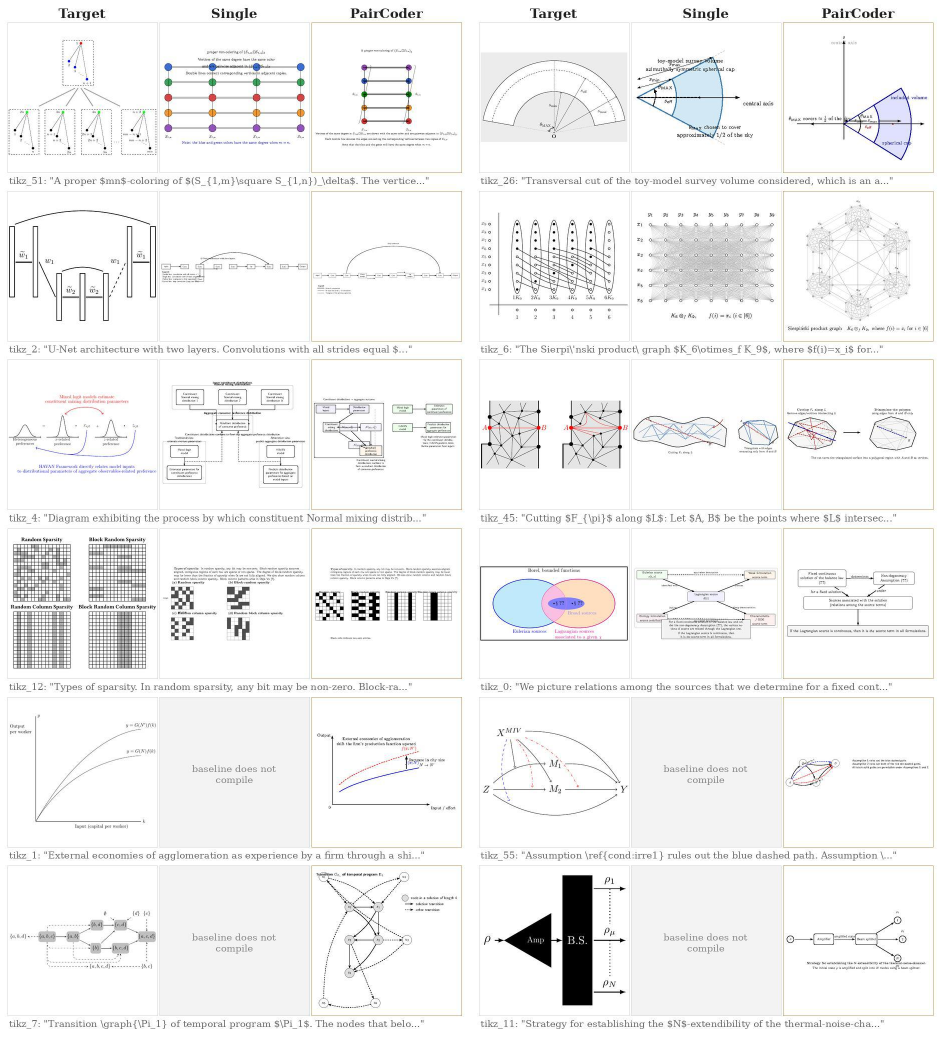}
    \captionof{figure}{DaTikZ qualitative gallery (12 distinct examples) demonstrating PairCoder's advantage; gray panels mark baselines that produce no artifact.}
    \label{fig:gal_datikz1}

    \vspace{0.5em}
    \centering\scriptsize\setlength{\tabcolsep}{4pt}\renewcommand{\arraystretch}{1.05}
    \begin{tabular}{lcccc}
\toprule
\textbf{Model} & compile rate $\uparrow$ & SSIM $\uparrow$ & CLIP $\uparrow$ & DINO $\uparrow$ \\
\midrule
gpt-5.4-mini & \cellcolor{improvebg}0.500 $\rightarrow$ \textbf{0.633} & \cellcolor{improvebg}0.215 $\rightarrow$ \textbf{0.256} & \cellcolor{improvebg}0.327 $\rightarrow$ \textbf{0.384} & \cellcolor{improvebg}0.303 $\rightarrow$ \textbf{0.338} \\
gpt-5.4 & \cellcolor{improvebg}0.717 $\rightarrow$ \textbf{0.967} & \cellcolor{improvebg}0.363 $\rightarrow$ \textbf{0.512} & \cellcolor{improvebg}0.566 $\rightarrow$ \textbf{0.770} & \cellcolor{improvebg}0.530 $\rightarrow$ \textbf{0.698} \\
gpt-5.5 & \cellcolor{improvebg}0.783 $\rightarrow$ \textbf{1.000} & \cellcolor{improvebg}0.394 $\rightarrow$ \textbf{0.485} & \cellcolor{improvebg}0.602 $\rightarrow$ \textbf{0.767} & \cellcolor{improvebg}0.554 $\rightarrow$ \textbf{0.730} \\
doubao-1.5-lite & \cellcolor{improvebg}0.550 $\rightarrow$ \textbf{0.850} & \cellcolor{improvebg}0.406 $\rightarrow$ \textbf{0.601} & \cellcolor{improvebg}0.391 $\rightarrow$ \textbf{0.621} & \cellcolor{improvebg}0.259 $\rightarrow$ \textbf{0.391} \\
doubao-seed-2.0-mini & \cellcolor{improvebg}0.567 $\rightarrow$ \textbf{0.800} & \cellcolor{improvebg}0.325 $\rightarrow$ \textbf{0.449} & \cellcolor{improvebg}0.441 $\rightarrow$ \textbf{0.606} & \cellcolor{improvebg}0.383 $\rightarrow$ \textbf{0.545} \\
deepseek-v3.2 & \cellcolor{improvebg}0.683 $\rightarrow$ \textbf{0.917} & \cellcolor{improvebg}0.365 $\rightarrow$ \textbf{0.544} & \cellcolor{improvebg}0.508 $\rightarrow$ \textbf{0.702} & \cellcolor{improvebg}0.465 $\rightarrow$ \textbf{0.596} \\
deepseek-v4-flash & \cellcolor{improvebg}0.633 $\rightarrow$ \textbf{0.733} & \cellcolor{improvebg}0.348 $\rightarrow$ \textbf{0.406} & \cellcolor{improvebg}0.505 $\rightarrow$ \textbf{0.578} & \cellcolor{improvebg}0.445 $\rightarrow$ \textbf{0.519} \\
\bottomrule
\end{tabular}
    \captionof{table}{DaTikZ: full measured metric suite for every applicable model (single model $\rightarrow$ \textbf{PairCoder}).}
    \label{tab:quant_datikz}
\end{figure*}

\begin{figure*}[p]
    \centering
    \includegraphics[width=\textwidth]{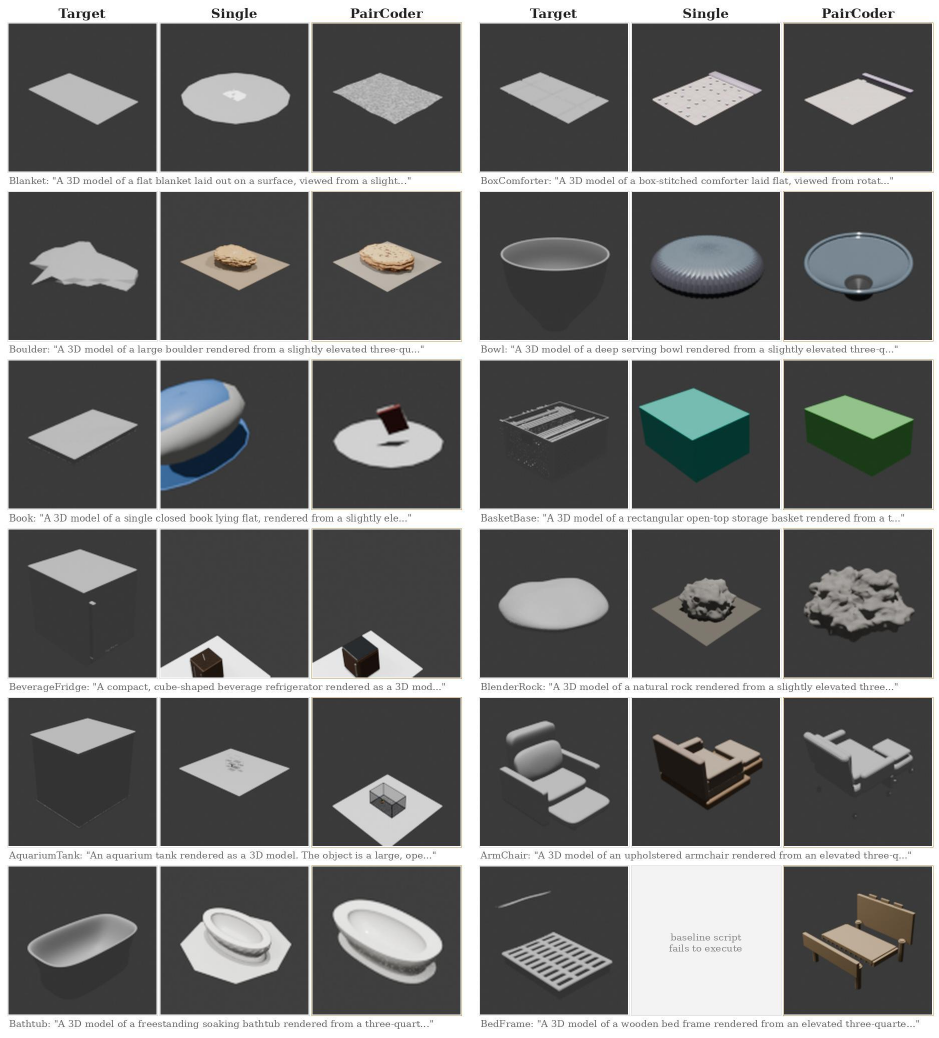}
    \captionof{figure}{3DCodeBench qualitative gallery (12 distinct examples) demonstrating PairCoder's advantage; gray panels mark baselines that produce no artifact.}
    \label{fig:gal_tdcb1}

    \vspace{0.5em}
    \centering\scriptsize\setlength{\tabcolsep}{4pt}\renewcommand{\arraystretch}{1.05}
    \begin{tabular}{lc}
\toprule
\textbf{Model} & executability $\uparrow$ \\
\midrule
gpt-5.4-mini & \cellcolor{improvebg}0.200 $\rightarrow$ \textbf{0.783} \\
gpt-5.4 & \cellcolor{improvebg}0.433 $\rightarrow$ \textbf{0.783} \\
gpt-5.5 & \cellcolor{improvebg}0.383 $\rightarrow$ \textbf{0.417} \\
doubao-1.5-lite & \cellcolor{improvebg}0.167 $\rightarrow$ \textbf{0.383} \\
doubao-seed-2.0-mini & \cellcolor{improvebg}0.200 $\rightarrow$ \textbf{0.533} \\
deepseek-v3.2 & \cellcolor{improvebg}0.333 $\rightarrow$ \textbf{0.633} \\
deepseek-v4-flash & 0.600 $\rightarrow$ 0.533 \\
\midrule
\multicolumn{2}{l}{\textit{Visual / geometric fidelity at \texttt{gpt-5.4-mini} (aggregate, n=60):}} \\
SigLIP-2 / DINO $\uparrow$ & \cellcolor{improvebg}.181/.101 $\rightarrow$ \textbf{.702/.376} \\
Chamfer (aggregate) $\downarrow$ & \cellcolor{improvebg}4.85 $\rightarrow$ \textbf{2.62} \\
\bottomrule
\end{tabular}
    \captionof{table}{3DCodeBench: full measured metric suite for every applicable model (single model $\rightarrow$ \textbf{PairCoder}).}
    \label{tab:quant_tdcb}
\end{figure*}

\begin{table*}[p]
\caption{Per-benchmark token accounting at \texttt{gpt-5.4-mini}: single model and PairCoder totals over the full task set, with the cost multiplier; the final row aggregates all runs across all models.}
\label{tab:quant_tokens}
\centering\scriptsize\setlength{\tabcolsep}{5pt}\renewcommand{\arraystretch}{1.12}
\begin{tabular}{lrrr}
\toprule
\textbf{Benchmark} & single model tokens & PairCoder tokens & multiplier \\
\midrule
RTLLM & 161,708 & 1,272,758 & 7.9$\times$ \\
BigCodeBench & 428,875 & 3,514,486 & 8.2$\times$ \\
DS-1000 & 550,911 & 1,995,002 & 3.6$\times$ \\
HumanEval-X JS & 445,887 & 2,575,842 & 5.8$\times$ \\
WebApp1K & 310,709 & 1,234,990 & 4.0$\times$ \\
DaTikZ & 260,271 & 1,802,890 & 6.9$\times$ \\
StarVector & 212,628 & 717,238 & 3.4$\times$ \\
ChartMimic & 226,356 & 1,341,183 & 5.9$\times$ \\
GenCAD-Code & 205,964 & 597,230 & 2.9$\times$ \\
Plot2Code & 216,997 & 2,004,708 & 9.2$\times$ \\
PandasPlotBench & 227,343 & 1,439,613 & 6.3$\times$ \\
\midrule
\textbf{All runs, all models} & 16,373,001 & 121,722,770 & 7.4$\times$ \\
\bottomrule
\end{tabular}
\end{table*}

\section{Cross-Model Qualitative Comparisons}
\label{app:crossmodel}
\begin{figure*}[p]
    \centering
    \includegraphics[width=\textwidth,height=0.92\textheight,keepaspectratio]{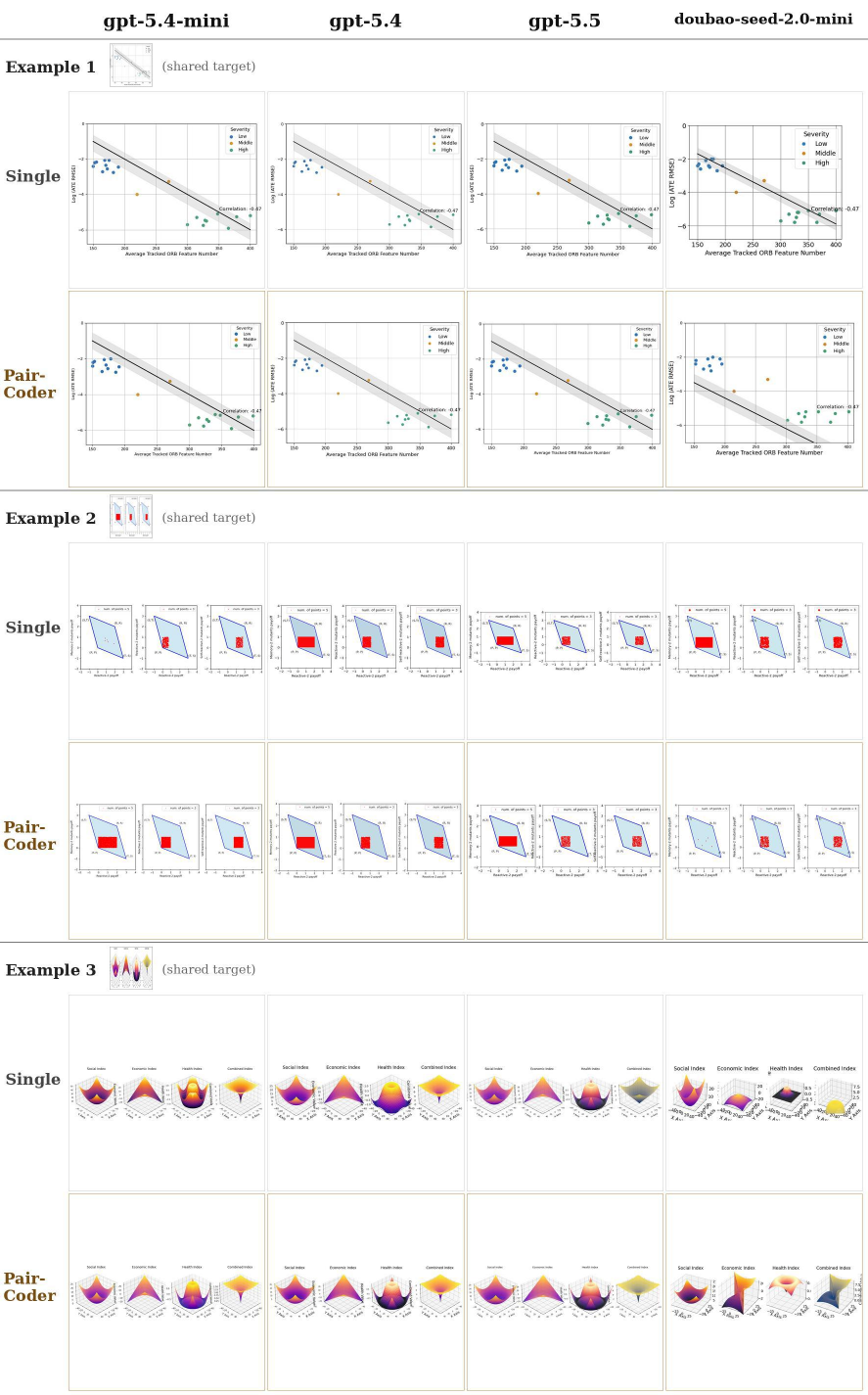}
    \caption{ChartMimic across four models and three shared tasks (single vs.\ PairCoder per cell).}
    \label{fig:xm_chart}
\end{figure*}

\begin{figure*}[p]
    \centering
    \includegraphics[width=\textwidth,height=0.92\textheight,keepaspectratio]{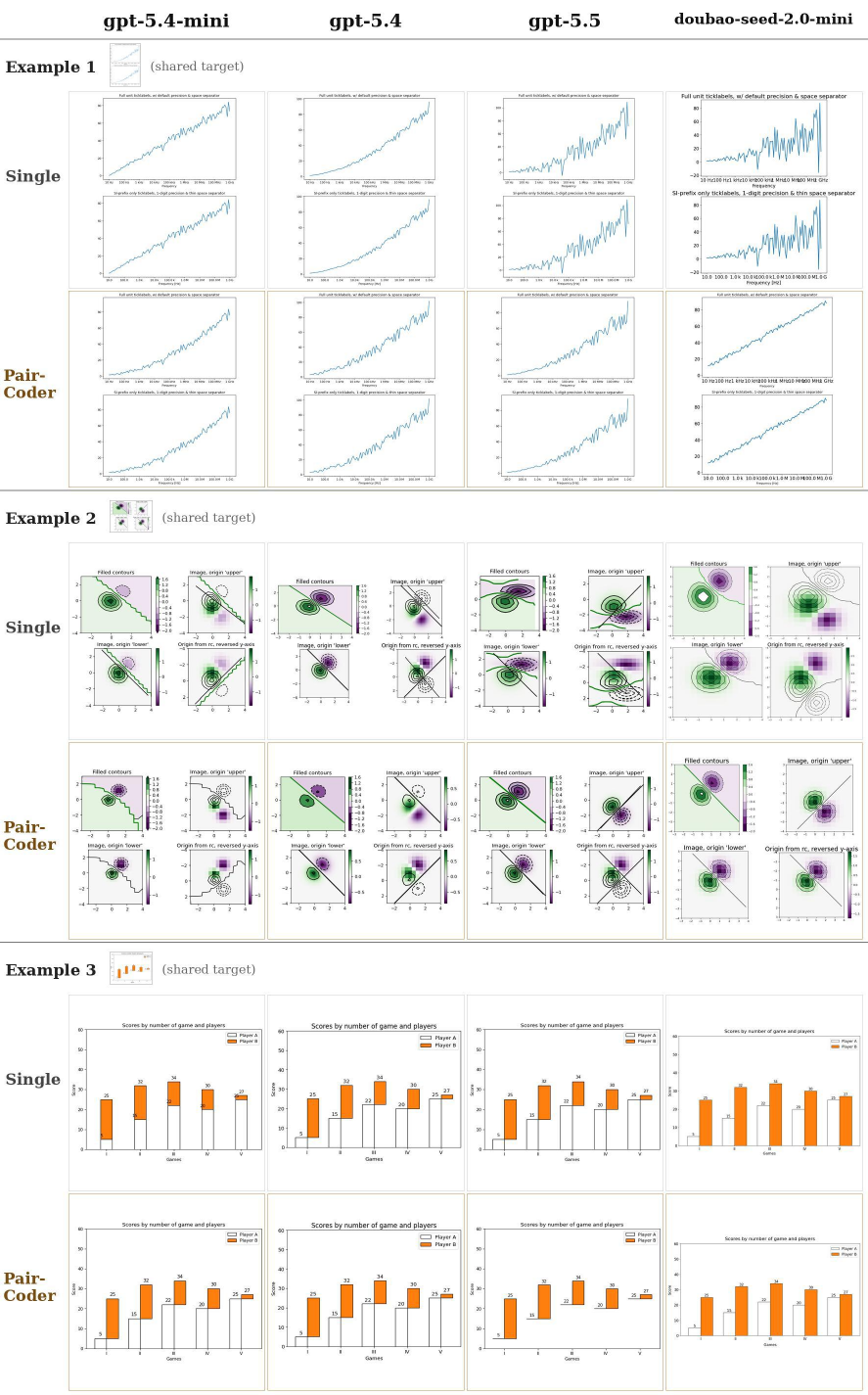}
    \caption{Plot2Code across four models and three shared tasks.}
    \label{fig:xm_p2c}
\end{figure*}

\begin{figure*}[p]
    \centering
    \includegraphics[width=\textwidth,height=0.92\textheight,keepaspectratio]{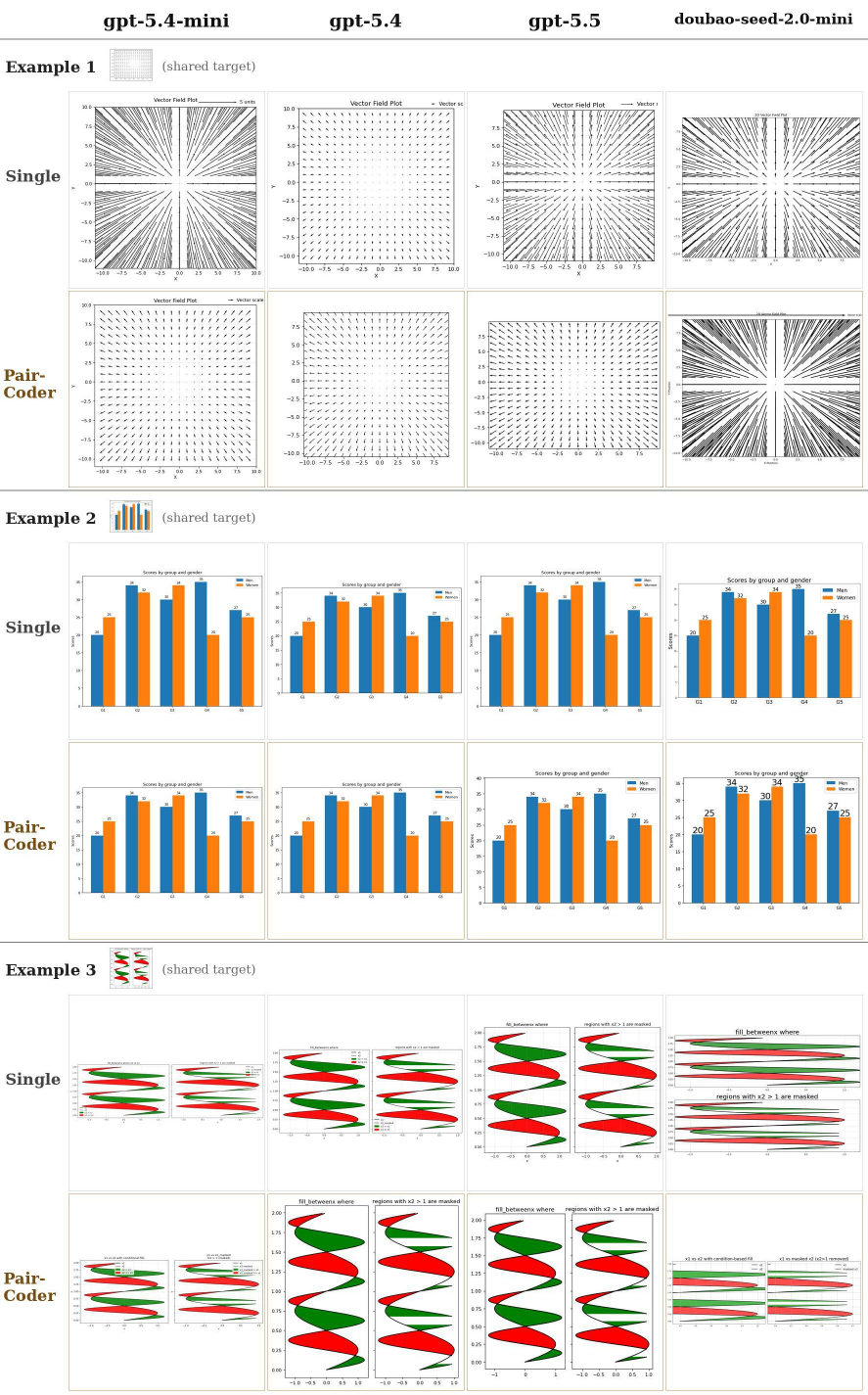}
    \caption{PandasPlotBench across four models and three shared tasks.}
    \label{fig:xm_ppb}
\end{figure*}

\begin{figure*}[p]
    \centering
    \includegraphics[width=\textwidth,height=0.92\textheight,keepaspectratio]{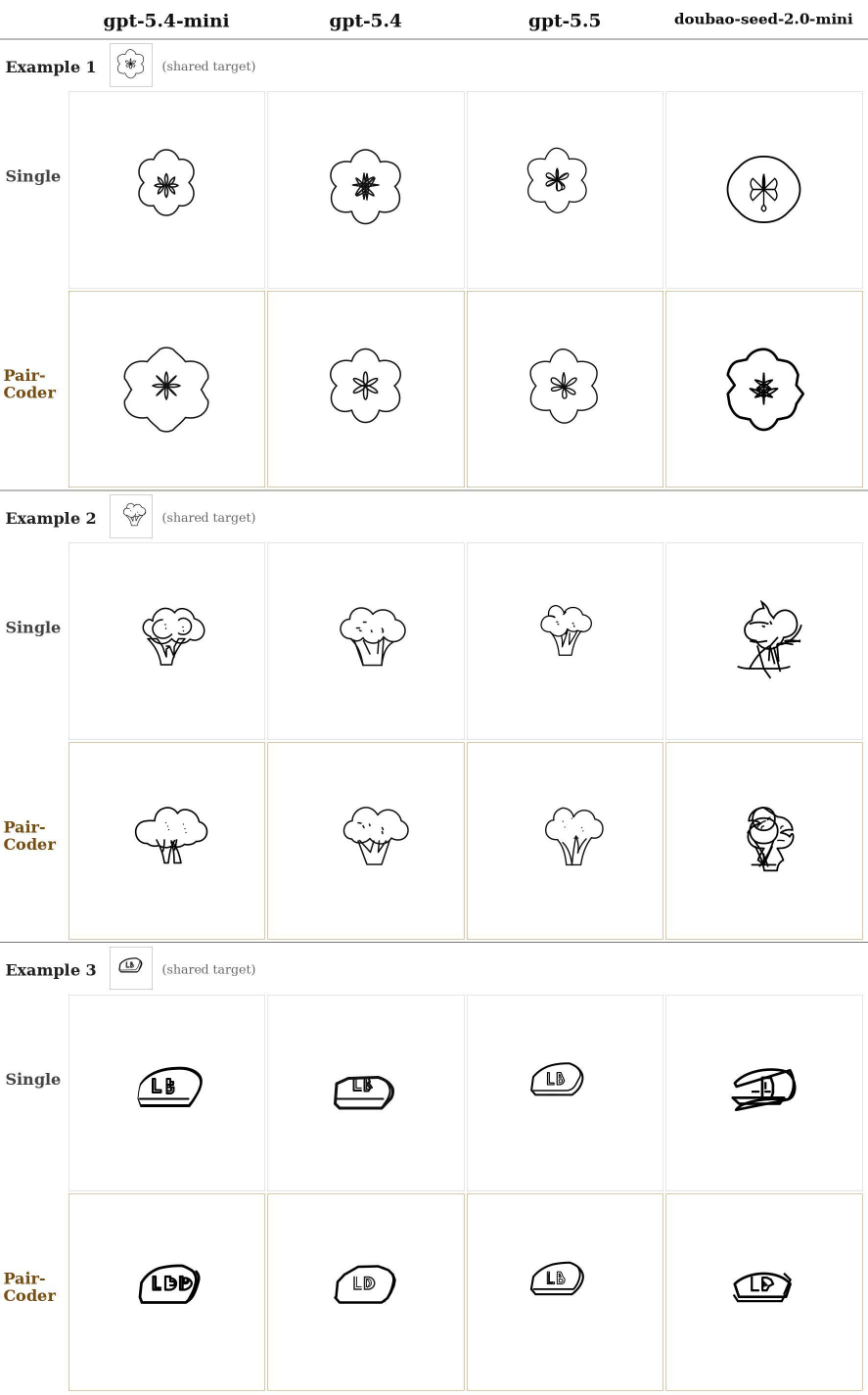}
    \caption{StarVector across four models and three shared tasks.}
    \label{fig:xm_svg}
\end{figure*}

\begin{figure*}[p]
    \centering
    \includegraphics[width=\textwidth,height=0.92\textheight,keepaspectratio]{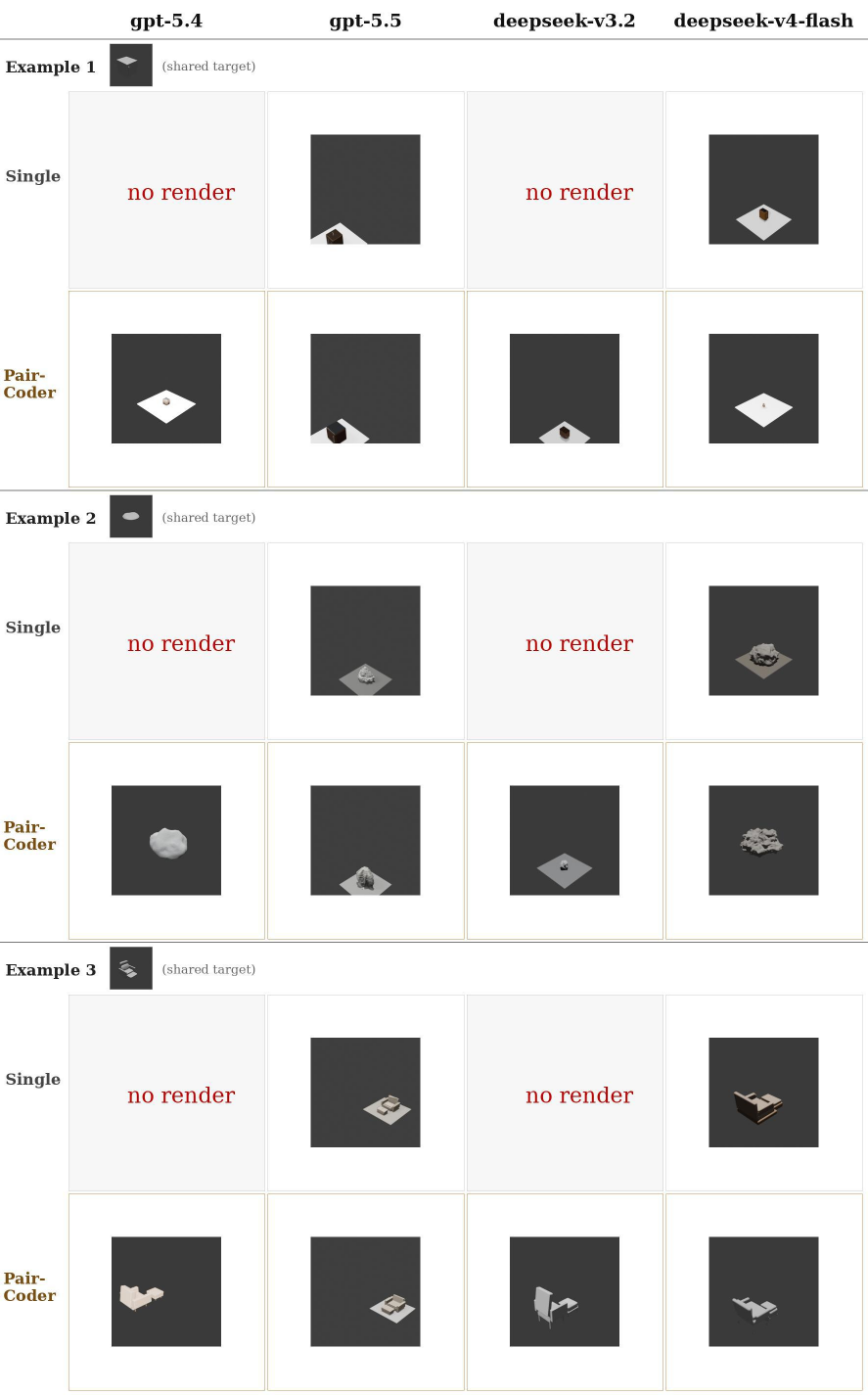}
    \caption{3DCodeBench across four models and three shared tasks; gray marks baselines that fail to render.}
    \label{fig:xm_tdcb}
\end{figure*}

The per-benchmark galleries above use \texttt{gpt-5.4-mini}. To show that the improvement is
not specific to one backbone, Figures~\ref{fig:xm_chart}--\ref{fig:xm_tdcb} hold the benchmark
and the task fixed and vary the base model: each page covers one benchmark across four models
(columns) and three shared tasks (stacked blocks, the shared target shown in each strip), with
the single model artifact above the PairCoder artifact for the same model and task. The chart,
plot, and vector panels use tasks where both arms render, so the comparison isolates quality;
the 3DCodeBench panel uses tasks where several backbones fail to render and PairCoder repairs
execution (gray ``no render'' marks the failure).


\section{P3D-Bench: Detailed Results}
\label{app:p3d}

\paragraph{Full geometry breakdown.}
Table~\ref{tab:p3d_detail} gives the full P3D-Bench geometry breakdown, adding the capped Chamfer
score (CD) to the F-score, normal-consistency (NC), and volumetric-IoU sub-metrics already reported
per model in Table~\ref{tab:main_grid}. All are the
worst-filled mean over the 400-case text-to-3D split (invalid programs score $0$, so the
metric jointly rewards executability and fidelity). PairCoder improves \emph{every}
sub-metric on \emph{every} model. The gains are largest exactly where the single-model
baseline leaves headroom: on \texttt{gpt-5.4} ($71.5\%$ valid) F-score@0.05 rises
$0.443\rightarrow0.619$, NC $0.435\rightarrow0.605$, and IoU $0.226\rightarrow0.334$, while
on the weak \texttt{doubao-1.5-lite} ($14.0\%$ valid) every sub-metric roughly triples.
Two facts isolate \emph{where} the gain comes from. First, on the cases the baseline already
compiles, the mean raw Chamfer distance (unnormalized, lower better) is essentially unchanged
(e.g.\ $0.0042\rightarrow0.0042$
on \texttt{doubao-seed-2.0-mini}, $0.0029\rightarrow0.0029$ on \texttt{gpt-5.4-mini}): PairCoder
does not perturb solids that already build. Second, the worst-filled sub-metrics nonetheless
rise sharply, because PairCoder converts a large fraction of \emph{non-executable} programs into
valid solids that recover the target's structure: the coverage gain, not a re-optimization of
already-valid geometry, drives the numbers, consistent with the conservative-review design.

\begin{table*}[t]
\centering
\caption{P3D-Bench text-to-3D \textbf{geometry} sub-metrics (single $\rightarrow$ \textbf{PairCoder},
400 cases, worst-filled means, higher better). CD is the capped-normalized Chamfer score
$1-\min(\mathrm{CD},\tau)/\tau$ ($\tau{=}0.01$).}
\label{tab:p3d_detail}
\small
\setlength{\tabcolsep}{3.5pt}
\begin{tabular}{lcccccc}
\toprule
Model & Validity $\uparrow$ & CD $\uparrow$ & F@0.05 $\uparrow$ & F@0.01 $\uparrow$ & NC $\uparrow$ & IoU $\uparrow$ \\
\midrule
\texttt{gpt-5.4-mini}          & 0.973$\rightarrow$\textbf{1.000} & 0.760$\rightarrow$\textbf{0.782} & 0.746$\rightarrow$\textbf{0.767} & 0.316$\rightarrow$\textbf{0.326} & 0.678$\rightarrow$\textbf{0.698} & 0.486$\rightarrow$\textbf{0.501} \\
\texttt{gpt-5.4}               & 0.715$\rightarrow$\textbf{0.990} & 0.453$\rightarrow$\textbf{0.623} & 0.443$\rightarrow$\textbf{0.619} & 0.153$\rightarrow$\textbf{0.215} & 0.435$\rightarrow$\textbf{0.605} & 0.226$\rightarrow$\textbf{0.334} \\
\texttt{gpt-5.5}               & 0.672$\rightarrow$\textbf{0.943} & 0.415$\rightarrow$\textbf{0.564} & 0.419$\rightarrow$\textbf{0.570} & 0.143$\rightarrow$\textbf{0.188} & 0.402$\rightarrow$\textbf{0.556} & 0.207$\rightarrow$\textbf{0.288} \\
\texttt{doubao-1.5-lite}       & 0.140$\rightarrow$\textbf{0.427} & 0.086$\rightarrow$\textbf{0.269} & 0.094$\rightarrow$\textbf{0.275} & 0.037$\rightarrow$\textbf{0.104} & 0.087$\rightarrow$\textbf{0.262} & 0.054$\rightarrow$\textbf{0.157} \\
\texttt{doubao-seed-2.0-mini}  & 0.812$\rightarrow$\textbf{0.990} & 0.542$\rightarrow$\textbf{0.641} & 0.537$\rightarrow$\textbf{0.633} & 0.196$\rightarrow$\textbf{0.230} & 0.504$\rightarrow$\textbf{0.609} & 0.302$\rightarrow$\textbf{0.360} \\
\texttt{deepseek-v3.2}         & 0.950$\rightarrow$\textbf{0.993} & 0.693$\rightarrow$\textbf{0.724} & 0.682$\rightarrow$\textbf{0.714} & 0.282$\rightarrow$\textbf{0.298} & 0.640$\rightarrow$\textbf{0.671} & 0.434$\rightarrow$\textbf{0.460} \\
\texttt{deepseek-v4-flash}     & 0.945$\rightarrow$\textbf{1.000} & 0.740$\rightarrow$\textbf{0.780} & 0.727$\rightarrow$\textbf{0.764} & 0.298$\rightarrow$\textbf{0.311} & 0.650$\rightarrow$\textbf{0.685} & 0.474$\rightarrow$\textbf{0.498} \\
\bottomrule
\end{tabular}
\end{table*}

\paragraph{Topology breakdown.}
Table~\ref{tab:p3d_topo} expands the P3D-Bench topology column of Table~\ref{tab:main_grid} into its
three official sub-metrics, again as worst-filled means: the fraction of solids with no open edge
(NoOE), and one minus the inverted-normal (InvN) and non-manifold-edge (NM) ratios; all lie in $[0,1]$
with higher better. The pattern mirrors geometry: PairCoder improves every topology sub-metric on every
model, most strongly where the baseline leaves headroom (\texttt{gpt-5.4} NoOE $0.695\rightarrow0.970$,
\texttt{doubao-1.5-lite} NoOE $0.130\rightarrow0.400$). Because a program must build a valid solid before
it can be watertight or manifold, these gains are again driven by converting non-executable programs into
well-formed solids rather than by repairing already-valid meshes.

\begin{table*}[t]
\centering
\caption{P3D-Bench text-to-3D \textbf{topology} sub-metrics (single $\rightarrow$ \textbf{PairCoder},
400 cases, worst-filled means, higher better): no-open-edge fraction (NoOE), and one minus the
inverted-normal (InvN) and non-manifold-edge (NM) ratios.}
\label{tab:p3d_topo}
\small
\setlength{\tabcolsep}{16pt}
\begin{tabular}{lccc}
\toprule
Model & NoOE $\uparrow$ & InvN $\uparrow$ & NM $\uparrow$ \\
\midrule
\texttt{gpt-5.4-mini}          & 0.963$\rightarrow$\textbf{0.990} & 0.973$\rightarrow$\textbf{1.000} & 0.968$\rightarrow$\textbf{0.996} \\
\texttt{gpt-5.4}               & 0.695$\rightarrow$\textbf{0.970} & 0.715$\rightarrow$\textbf{0.990} & 0.714$\rightarrow$\textbf{0.988} \\
\texttt{gpt-5.5}               & 0.662$\rightarrow$\textbf{0.930} & 0.672$\rightarrow$\textbf{0.943} & 0.672$\rightarrow$\textbf{0.940} \\
\texttt{doubao-1.5-lite}       & 0.130$\rightarrow$\textbf{0.400} & 0.140$\rightarrow$\textbf{0.427} & 0.139$\rightarrow$\textbf{0.419} \\
\texttt{doubao-seed-2.0-mini}  & 0.800$\rightarrow$\textbf{0.978} & 0.812$\rightarrow$\textbf{0.990} & 0.811$\rightarrow$\textbf{0.988} \\
\texttt{deepseek-v3.2}         & 0.940$\rightarrow$\textbf{0.983} & 0.950$\rightarrow$\textbf{0.993} & 0.949$\rightarrow$\textbf{0.992} \\
\texttt{deepseek-v4-flash}     & 0.945$\rightarrow$\textbf{1.000} & 0.945$\rightarrow$\textbf{1.000} & 0.939$\rightarrow$\textbf{0.994} \\
\bottomrule
\end{tabular}
\end{table*}

\begin{figure*}[p]
    \centering
    \includegraphics[width=\textwidth]{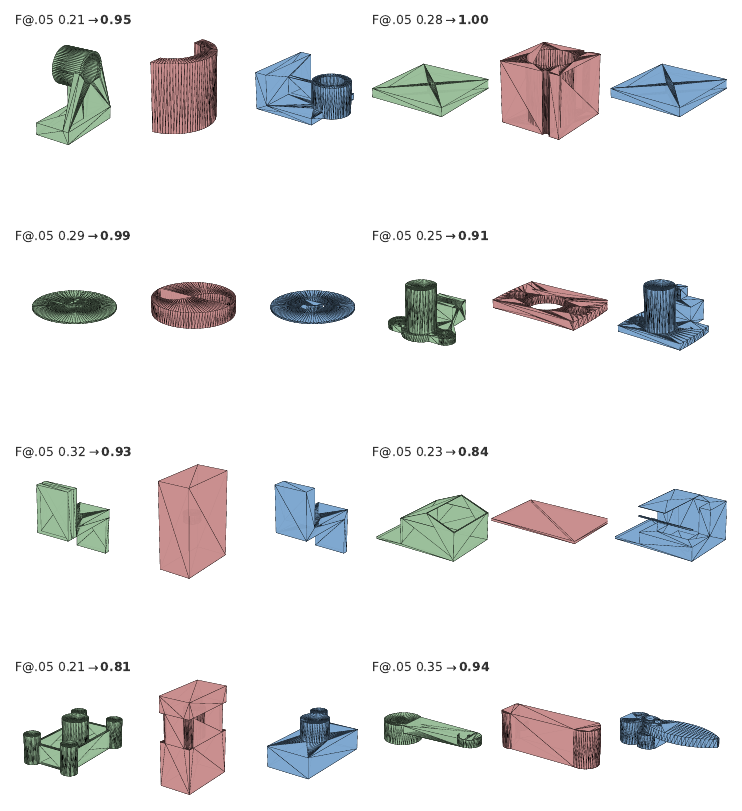}
    \caption{P3D-Bench \textbf{geometry-refinement} gallery (\texttt{gpt-5.4}): ground truth (green),
    single model (pink; valid but geometrically poor), PairCoder (blue); per-case F-score@0.05 annotated.}
    \label{fig:p3d_improved}
\end{figure*}

\paragraph{Qualitative examples: geometry refinement.}
On P3D-Bench text-to-3D the compile-grounded PairCoder of Table~\ref{tab:main_grid} gains by
\emph{coverage}: it repairs non-executable programs into valid solids and leaves already-compiling
programs unchanged, so on the common set the two solids are identical. We additionally evaluate a
vision-grounded geometry-refinement variant: when the single-model program compiles to a valid but
geometrically poor solid, the Navigator renders the solid and reviews it against the specification,
and the Driver refines the parametric program (kept only if it still compiles). Across a sample of
$40$ mid-quality valid solids ($0.2<\text{F@}0.05<0.8$) on \texttt{gpt-5.4}, this refinement raises
F-score@0.05 on $29$ and regresses on $3$ (the kept-if-compiles safeguard bounds regressions), for a
mean gain of $+0.26$; it is thus a genuine effect rather than a cherry-pick.
Figure~\ref{fig:p3d_improved} shows cases where both the single model and this variant build valid
solids, but the refined one is much closer to the target: each triplet renders the ground-truth
solid (green), the single-model solid (pink; valid but wrong), and the PairCoder-refined solid
(blue), with the per-case F-score@0.05 improvement annotated.

\paragraph{A second Text-to-3D format: OpenSCAD.}
To test whether the P3D-Bench gain is specific to the minimal-JSON format, we repeat the
400-case Text-to-3D experiment with OpenSCAD, a constructive-solid scripting language, on the
three models for which both formats were run. Table~\ref{tab:p3d_openscad} reports validity, the
fraction of programs that compile to a valid solid, for both formats side by side. Two
observations stand out. First, PairCoder drives OpenSCAD validity to a perfect $1.000$ on all
three models, repairing every non-compiling program. Second, the single-model baseline is
markedly more fluent in OpenSCAD than in the minimal-JSON format (for example \texttt{gpt-5.4}
baseline validity $0.715\rightarrow0.932$), so the coverage headroom PairCoder recovers is
correspondingly smaller: its OpenSCAD validity gain is $+6.8$ points on \texttt{gpt-5.4} versus
$+27.5$ points in minimal-JSON. This is the same headroom-tracking pattern seen across the rest
of the grid. The gain concentrates where the baseline leaves room; OpenSCAD's higher baseline
fluency compresses it, while PairCoder still closes the remaining gap to perfect executability
in both formats.

\begin{table*}[t]
\centering
\caption{P3D-Bench text-to-3D validity across two program formats (single $\rightarrow$
\textbf{PairCoder}, 400 cases, thinking off).}
\label{tab:p3d_openscad}
\small
\setlength{\tabcolsep}{18pt}
\begin{tabular}{lcc}
\toprule
 & \multicolumn{2}{c}{Validity $\uparrow$ (single $\rightarrow$ \textbf{PC})} \\
\cmidrule(lr){2-3}
Model & minimal-JSON & OpenSCAD \\
\midrule
\texttt{gpt-5.4}              & 0.715$\rightarrow$\textbf{0.990} & 0.932$\rightarrow$\textbf{1.000} \\
\texttt{doubao-seed-2.0-mini} & 0.812$\rightarrow$\textbf{0.990} & 0.925$\rightarrow$\textbf{1.000} \\
\texttt{deepseek-v3.2}        & 0.950$\rightarrow$\textbf{0.993} & 0.990$\rightarrow$\textbf{1.000} \\
\bottomrule
\end{tabular}
\end{table*}

\paragraph{Image-to-3D and assembly-3D demonstration.}
The other two P3D-Bench tracks, image-to-3D and assembly-3D, condition on a rendered image of the
target (assembly-3D also on part annotations) and require a vision-capable model; their full
evaluation sets build on the licensed Fusion~360 Gallery. We run the three in-repo demo cases per
track with \texttt{gpt-5.4} across the CadQuery and Three.js program formats (OpenSCAD is already
evaluated at scale on text-to-3D, Table~\ref{tab:p3d_openscad}), scoring the same non-VLM buckets
as text-to-3D. Table~\ref{tab:p3d_demo} reports the result. In CadQuery, PairCoder repairs the one
non-compiling image-to-3D case ($0.667\rightarrow1.000$ valid) and recovers a valid assembly-3D
solid from a baseline that builds none ($0\rightarrow0.333$), via a sibling candidate. In Three.js,
\texttt{gpt-5.4} already compiles all three image-to-3D cases, so PairCoder ties, consistent with
the headroom pattern seen across the grid. The three-case sample is illustrative rather than
conclusive, but it shows the compiler-grounded mechanism reaches the image-conditioned tracks and a
web-rendering representation, spanning all four P3D-Bench program formats together with text-to-3D.

\begin{table*}[t]
\centering
\caption{P3D-Bench image-to-3D and assembly-3D \textbf{three-case demonstration}
(single $\rightarrow$ \textbf{PairCoder}, \texttt{gpt-5.4}) in the CadQuery and Three.js formats;
full evaluation requires the licensed Fusion~360 Gallery.}
\label{tab:p3d_demo}
\small
\setlength{\tabcolsep}{10pt}
\begin{tabular}{llccc}
\toprule
Track & Format & Validity $\uparrow$ & Geometry $\uparrow$ & Topology $\uparrow$ \\
\midrule
Image-to-3D & CadQuery & 0.667$\rightarrow$\textbf{1.000} & 0.100$\rightarrow$\textbf{0.320} & 0.667$\rightarrow$\textbf{1.000} \\
Image-to-3D & Three.js & 1.000$\rightarrow$1.000 & 0.417$\rightarrow$0.417 & 0.778$\rightarrow$0.778 \\
Assembly-3D & CadQuery & 0.000$\rightarrow$\textbf{0.333} & 0.000$\rightarrow$\textbf{0.068} & 0.000$\rightarrow$\textbf{0.333} \\
\bottomrule
\end{tabular}
\end{table*}

\section{Reproducibility Notes}
\label{app:repro}
For every benchmark we use the official task set and scorer where released (ChartMimic,
Plot2Code, PandasPlotBench, StarVector, DaTikZ, 3DCodeBench, GenCAD-Code, VerilogEval, RTLLM,
WebApp1K, LiveCodeBench, BigCodeBench, DS-1000, HumanEval-X), identical prompts and code
extraction in both arms, and provider default sampling with thinking disabled. Visual metrics
are computed with SSIM at 256$\times$256 and the public CLIP, DINOv2, and SigLIP-2
checkpoints; Chamfer distance is computed on normalized surface samples, and aggregate
variants assign non executing generations the worst observed score so that executability
cannot be traded for similarity. Token counts are taken from provider usage fields and logged
separately for the baseline arm and the PairCoder arm.

\end{document}